\documentclass[journal]{IEEEtran}

\usepackage[utf8]{inputenc}
\usepackage{siunitx}
\usepackage[colorlinks=true, allcolors=blue]{hyperref}
\usepackage{amsmath}
\usepackage{amssymb}
\usepackage{amsfonts}
\usepackage{soul} %ONLY FOR THE DRAFT VERSION, REMOVE AFTERWARDS
\usepackage{subcaption}
\usepackage{stfloats}
\usepackage{booktabs}
\usepackage{enumitem}
\usepackage{makecell}

\usepackage{tikz,xcolor}
\usepackage{array}
\usepackage{footnote}
\makesavenoteenv{tabular}

\usepackage{algorithm}  
\usepackage{algpseudocode}  
\usepackage{amsmath}

\definecolor{lime}{HTML}{A6CE39}
\DeclareRobustCommand{\orcidicon}
{
    \begin{tikzpicture}
    \draw[lime, fill=lime] (0,0) circle [radius=0.16] 
    node[white] {{\fontfamily{qag}\selectfont \tiny ID}};    \draw[white, fill=white] (-0.0625,0.095) circle [radius=0.007];    
    \end{tikzpicture}
    \hspace{0mm}}
\foreach \x in {A, ..., Z}{%
    \expandafter\xdef\csname orcid\x\endcsname{\noexpand\href{https://orcid.org/\csname orcidauthor\x\endcsname}{\noexpand\orcidicon}}
}

\setlist[enumerate]{itemsep = 0pt, parsep = 0pt, topsep = 0pt} % enumerate spacing setting
\setlist[itemize]{itemsep = 0pt, parsep = 0pt, topsep = 0pt} % itemize spacing setting

\begin{document}  

\title{Research on Reliable and Safe Occupancy Grid Prediction in Underground Parking Lots}

% https://www.programmersought.com/article/12636730965/
\author{
Jiaqi Luo \vspace{-20pt}
\thanks{Jiaqi Luo is with the Department of Computer Science and Technology, Wuhan University of Science and Technology, Wuhan, China}
}

\markboth{Journal of \LaTeX\ Class Files,~Vol.
% ~14, No.~8, August~2015
}%
{Shell \MakeLowercase{\textit{et al.}}: Bare Demo of IEEEtran.cls for IEEE Journals}

% Make the title area
\maketitle

% As a general rule, do not put math, special symbols, or citations
\begin{abstract}
Against the backdrop of advancing science and technology, autonomous vehicle technology has emerged as a focal point of intense scrutiny within the academic community. 
Nevertheless, the challenge persists in guaranteeing the safety and reliability of this technology when navigating intricate scenarios.

While a substantial portion of autonomous driving research is dedicated to testing in open-air environments, such as urban roads and highways, where the myriad variables at play are meticulously examined, enclosed indoor spaces like underground parking lots have, to a significant extent, been overlooked in the scholarly discourse. 
This discrepancy highlights a gap in understanding the unique challenges these confined settings pose for autonomous navigation systems.

This study tackles indoor autonomous driving, particularly in overlooked spaces like underground parking lots.
Using CARLA's simulation platform, a realistic parking model is created for data gathering.
An occupancy grid network then processes this data to predict vehicle paths and obstacles, enhancing the system's perception in complex indoor environments \cite{lan2023end}. 
Ultimately, this strategy improves safety in autonomous parking operations.

The paper meticulously evaluates the model's predictive capabilities, validating its efficacy in the context of underground parking. 
Our findings confirm that the proposed strategy successfully enhances autonomous vehicle performance in these complex indoor settings.
It equips autonomous systems with improved adaptation to underground lots, reinforcing safety measures and dependability.
This work paves the way for future advancements and applications by addressing the research shortfall concerning indoor parking environments, serving as a pivotal reference point.

\end{abstract}


% \fi
\IEEEpeerreviewmaketitle

%%%%%%%%%%%%%%%%%%%%%%%%%%%%%%%%%%%%%%%%%%%%%%
\section{Introduction}
%%%%%%%%%%%%%%%%%%%%%%%%%%%%%%%%%%%%%%%%%%%%%%
Autonomous driving has emerged as a crucial area of exploration within the automotive realm, with perception systems playing a vital role in guiding vehicles' actions and reactions.
These systems act as the eyes of autonomous cars, deciphering the environment to inform navigation decisions and ensure safe maneuvers.

Researchers integrated Bird's Eye View (BEV) \cite{liu2023bevfusion} perception into the mix to further enhance the safety and dependability of autonomous vehicles.
This innovative perspective offers a comprehensive overhead visual of the vehicle's surroundings, synthesizing data from diverse sensors.
BEV perception enables vehicles to meticulously discern roads, vehicles, pedestrians, and any potential obstructions by presenting a top-down view of the world, thereby significantly upgrading the accuracy and responsiveness of the autonomous driving system.

The essence of BEV perception revolves around consolidating sensory inputs from various detectors into a single, cohesive framework—effectively translating them into a bird's-eye-view format.
This transformation allows for the fusion and analysis of environmental features under a unified spatial context, forming the basis for sophisticated decision-making processes.
In implementing multi-sensor strategies within BEV perception, two principal methodologies prevail:
\begin{enumerate}[leftmargin=*]
    \item Multi-modal Fusion Approach

    This methodology harnesses the synergy of LiDAR and millimeter-wave radar, among other sensors. 
    The system acquires robust and precise data on both stationary and moving elements in the vehicle's vicinity by leveraging the active sensing capabilities of these devices, which emit signals and analyze their reflections.
    This comprehensive dataset, enriched with depth and positional accuracy, forms a cornerstone for numerous autonomous driving systems, reflecting the industry's widespread endorsement.
    \item Vision-Based Scheme

    Conversely, this approach banks on passive visual sensors, primarily cameras, to perceive the environment.
    Vision-based systems derive understanding solely from ambient light, interpreting visual cues to discern objects and navigate surroundings without emitting any probing signals.
    This method demands advanced image processing algorithms to compensate for the lack of direct distance measurement, making it a compelling alternative or supplementary pathway in autonomous vehicle design.
\end{enumerate}

Taking Tesla's FSD vision-based system as an example, multiple cameras are placed around the car body. 
The space around the car is represented as a two-dimensional grid in BEV coordinates.
BEV perception fuses features from multiple images into the corresponding two-dimensional BEV grid to provide a global view, and each grid corresponds to a part of the input image. 
The vision-based scheme is more technically difficult than the "multi-mode fusion" scheme that relies on lidar.
However, the cost is lower after successful research and development, and the production cost can be greatly reduced. 
As research progresses, vision-based approaches steadily increase in precision, demonstrating that they can approximate the perception performance of "multi-modal fusion" systems while maintaining a significantly lower cost profile \cite{lan2022vision}. 
This narrowing gap in effectiveness, coupled with reduced expenses, underscores the growing appeal and practicality of vision-based BEV perception in the autonomous driving landscape, fostering an environment where cost-efficient yet highly capable autonomous systems are increasingly feasible.
Therefore, BEV visual perception has set off a wave of research in academia and industry.

As one of the core technologies of BEV's visual perception, the occupying grid network is a 3D reconstruction method based on deep learning that enables the understanding of 3D space by dividing it into fixed-size voxels and predicting whether each voxel grid is occupied and which targets categories it may contain.
Compared with the two-dimensional grid described by BEV perception, the occupying grid network can describe the three-dimensional grid.
The occupancy grid network can predict whether a given spot is occupied even without identifying the specific target type.
It overcomes the limitation of treating undetermined objects as non-obstacles and enables a superior comprehension and handling of three-dimensional space.
Consequently, the system can execute more precise and swift maneuvers, demonstrating the potency of BEV perception in achieving heightened operational accuracy and efficiency.
% Significantly enhance your ability to understand your surroundings.

Indeed, a considerable portion of the existing research surrounding occupancy grid networks has predominantly revolved around outdoor environments.
There are two main reasons: firstly, there are more pressing and immediate demands for autonomous driving solutions in outdoor settings like highways and city streets. 
Secondly, it stems from the scarcity of autonomous driving datasets that encompass indoor scenarios, thereby inadvertently limiting the scope of studies in this domain.
The development and optimization of occupancy grid networks for indoor spaces, like underground parking lots, represents a promising yet relatively untapped avenue for future research. 
Therefore, this paper chooses the SurroundOcc \cite{wei2023surroundocc} approach based on occupying grid networks to perform perception tasks in underground parking lots in the current research state.

\begin{figure*}
    \centering
    \includegraphics[width=0.98\linewidth,trim={200 100 420 200},clip]{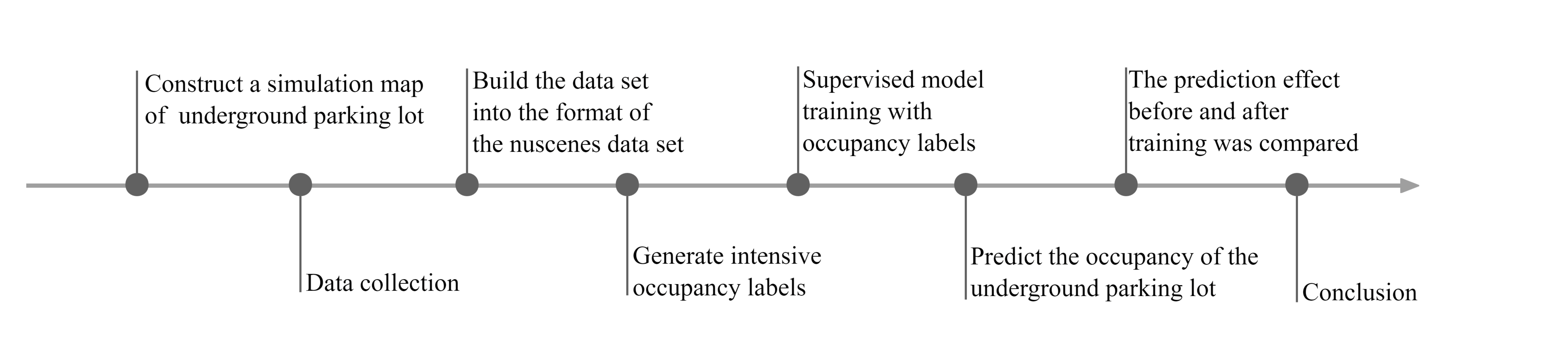}
    \caption{This figure shows the timeline of this work. 
    First, we gathered extensive data reflective of real-world parking dynamics based on the creation of a meticulous simulated underground parking lot environment within CARLA.
    The collected data is then used to generate intensive occupancy labels.
    Next, leveraging these meticulously crafted occupancy labels, we embarked on a supervised learning process for the "SurroundOcc" model.
    Finally, the prediction results are evaluated and summarized}
    \label{fig:timeline}
\end{figure*}

This study chooses the data set collected in self-built underground parking lot to carry out the research on this scenario in order to solve the scarcity of data sets.
Since there is no physical vehicle to collect parking lot data in the real world, this paper uses CARLA simulator to construct the scene of underground parking lot and collect data.
The SurroundOcc system has demonstrated promising performance using the nuScenes dataset, which is primarily geared toward outdoor and street-level scenarios. 
However, attempting to apply pre-trained models from SurroundOcc \cite{wei2023surroundocc} to an underground parking lot dataset directly revealed a major shortcoming. 
The model, trained primarily on outdoor scenes, failed to generalize to the distinct indoor environment of parking garages effectively \cite{liu2022towards}. 
This underscores the necessity for specialized training and fine-tuning of models to cater to the unique characteristics and challenges presented by indoor parking scenarios.
This paper proceeds by employing a dataset comprising detailed occupancy labels derived from underground parking scenarios to train the "SurroundOcc" model to address the challenge. 
Then the model is endowed with the capability to discern and predict occupancy within the complex parking environment meticulously.
This tailored training regimen ensures that the model is optimized for the specificities of parking scenes, ultimately enabling it to perform occupancy prediction accurately.
 \autoref{fig:timeline} shows the timeline of this work.

Through the scheme of this study, the automatic driving technology can be better applied in the underground parking lot, and the perception ability and safety of the automatic driving system in the underground parking lot can be improved \cite{yi2024key}.
In addition, it fills the gap in the research of automatic driving in underground parking lot scenarios and provides a reference for realizing automatic driving in a wider range of scenarios.

The main contributions of this study are as follows:
\begin{enumerate}
    \item Data Collection
    
    Acknowledging the scarcity of underground parking data in existing public datasets, this study devises a systematic method to gather and create a specialized dataset for enhancing autonomous vehicle performance in indoor parking environments.
    The specific process is shown in the \autoref{fig:data_collect}.
    \item SurroundOcc Method

   This section focuses on understanding the design of the SurroundOcc model and applying our collected dataset to its training.
   Central to this process are two key steps: first, creating a detailed occupancy ground truth, and second, outlining the architecture and workings of the SurroundOcc model.

\end{enumerate}
This paper will introduce detailed methods for the above two problems. \autoref{fig:method} shows the process of the main methods.
%%%%%%%%%%%%%%%%%%%%%%%%%%%%%%%%%%%%%%%%%%%%%%
\section{Related Work}
\label{sec:related_work}
%%%%%%%%%%%%%%%%%%%%%%%%%%%%%%%%%%%%%%%%%%%%%%
In the realm of Bird's-Eye-View (BEV) perception, vision-based approaches have garnered significant interest within academic circles. 
They stand out due to their potential to decrease both the financial outlay and intricacy associated with LiDAR-driven 3D perception systems, while also offering enhanced versatility in application. 
This makes them an intriguing avenue for further examination. 
Specifically, among vision-based BEV methodologies, two frameworks have risen to prominence: LSS and Transformer networks.

In the LSS (Lift, splat, shooting) \cite{philion2020lift} scheme, the camera projects the real world onto the image plane. 
It is a process of 3d to 2d, which will lose the depth. 
The purpose of lift is to restore the depth of each pixel in the image, lifting the image from a 2d plane to a 3d space.
Following the lift operation in the LSS scheme, two distinct 3D point clouds are generated:
\begin{enumerate}[leftmargin=*]
    \item Cone Point Cloud

    This cloud precisely locates each point within the autonomous vehicle's coordinate framework, effectively mapping their real-world positions in 3D space.
    \item Context Feature Point Cloud

    Complementarily, this cloud encapsulates the contextual attributes affiliated with every point such as color, texture, or object characteristics, enriching the spatial information with descriptive details.
\end{enumerate}

The purpose of splat is to project context features into the BEV grid and construct BEV features.
In general, it can be divided into two aspects: Use 2D features to construct depth information and lift 2D features into 3D space. 
Encode 2D features into 3D space through 3D to 2D projection mapping.
The shoot component, unrelated to the perception process, is omitted from this discussion.

Inspired by LSS, more excellent BEV perception algorithms have subsequently appeared, such as BEVDepth \cite{li2023bevdepth}, which utilizes the point cloud of lidar on the basis of LSS to supervise the predicted depth and make the predicted distance closer to the real value. 
BEVDet \cite{huang2022bevdet4d} innovatively unified ring view detection into the BEV space, providing a good template for subsequent work.
BEVFusion \cite{liu2023bevfusion} optimizes BEV pooling on the basis of LSS and BEVDet, increasing the latency of attempted conversions by 40 times.

In the scheme based on Transformer \cite{han2021transformer}, features in the BEV space are first initialized, and then a multi-layer Transformer is used to interact with each image feature to obtain corresponding BEV features.
A representative method is DETR3D \cite{wang2022detr3d}, which operates prediction directly in 3D space. 
It extracts 2D features from multiple camera images and then uses a sparse set of 3D object queries to index these 2D features to 3D locations, associating them with multi-view images via a camera transformation matrix \cite{xu2019online}. 
BEVFormer \cite{li2022bevformer} proposed a spatial-temporal transformer that explores temporal and spatial characteristics through the pre-defined BEV queris sampled on grid.

BEV perception is a two-dimensional representation that shows the environment from a bird's eye view (i.e., from above).
Occupancy grid network, on the other hand, is a three-dimensional representation that divides the environment into a series of cubes (or voxels) and assigns each voxel a value indicating whether that voxel is occupied by an object.
Occupancy Networks \cite{mescheder2019occupancy} is the first work of the series. 
The author proposes Occupancy Network, a new 3D reconstruction method based on deep
learning.
On this basis, INRIA proposed MonoScene \cite{cao2022monoscene}, which is the first paper that does not rely on depth and radar sensors as input signals.
It can support indoor and outdoor scene prediction with only a single frame image as input.

TPVFormer \cite{huang2023tri} of Tsinghua University proposed a three-perspective (TPV) representation. 
Each point in the three-dimensional space is modeled by adding its projected features across the three planes.
A TPV encoder based on a Transformer (TPVFormer) is proposed to obtain TPV features effectively in order to upgrade image features to 3D TPV space.
In subsequent work, Tsinghua continued to introduce SurroundOcc \cite{wei2023surroundocc} to predict 3D occupancy of multi-camera images.
Firstly, multi-scale features are extracted for each image, and spatial 2D-3D attention is used to raise it to three-dimensional space. 
Three-dimensional convolution is then applied to progressively up-sample the volume features and supervision is applied on multiple layers. 
In this method, a pipeline is also designed to generate intensive ground truth of occupancy, which can save a lot of manpower.

In the same series of methods, OccFormer \cite{zhang2023occformer} realized the long-range, dynamic, and efficient encoding of 3D voxel features generated by the camera so as to deal with the semantic occupancy prediction of 3D volume effectively.

\autoref{tab:camera} shows a comparison of the current approaches to Occupancy Networks
\begin{table*}[!ht]  \centering \small
\setlength\tabcolsep{1pt} \renewcommand{\arraystretch}{1.0}
  \caption{comparison of the current approaches to Occupancy Networks}
  \label{tab:occ}
  \begin{tabular}{c|p{5.1cm}|p{5.1cm}|p{5.1cm}}   \toprule
    \makebox[2.5cm][c]{occ method} & \makebox[5cm][c]{feature} & \makebox[5cm][c]{advantage} & \makebox[5cm][c]{disadvantage}  \\   \midrule
    MonoScene \cite{cao2022monoscene}& Predict 3D occupancy directly from monocular images &  radar-independent  &  Intractable occlusion \\   \midrule
    OccFormer \cite{zhang2023occformer}&  Dual-path converter network processing footprint & The process is interpretive &  Requires a lot of computing resources \\  \midrule 
    VoxFormer \cite{li2023voxformer} & Two-stage design predicts occupancy
    &The memory usage is greatly reduced &  Performance is affected by depth estimation\\ \midrule 
    TPVFormer \cite{huang2023tri}& Three view (TPV) notation & Sparse point cloud monitoring can still effectively predict voxel occupancy & Large amounts of annotated data are required \\    \midrule
    SurroundOcc \cite{wei2023surroundocc}& A method to generate dense occupancy truth value is proposed &  Able to fully perceive 3D scenes &  High computational cost\\    \bottomrule
  \end{tabular}
\end{table*}

In this study, the SurroundOcc method is selected for modeling and predicting underground parking lots, primarily due to the following reasons:
\begin{enumerate}[leftmargin=*]
    \item The paucity of semantic categories within underground parking scenarios, predominantly comprising obstacles like vehicles and walls, presents a unique challenge. 
    However, SurroundOcc's capacity for 3D panoramic perception and its adeptness at predicting truncated, irregularly shaped objects with ambiguous semantics renders it exceptionally capable in managing the diverse array of elements typically encountered in parking lots, including vehicles, pillars, and walls.
    \item In this article, the pre-training model provided by SurroundOcc was completely invalidated in underground car park scenarios. 
    However, SurroundOcc is a deep learning-based approach, so the models in this study could be trained on parking lot data to adapt to underground parking environments.
    \item SurroundOcc designs a process to generate dense occupy labels, which saves significant labour and time costs.
\end{enumerate}
%%%%%%%%%%%%%%%%%%%%%%%%%%%%%%%%%%%%%%%%%%%%%%
\section{Methodology}
\label{sec:methodology}
%%%%%%%%%%%%%%%%%%%%%%%%%%%%%%%%%%%%%%%%%%%%%%

\begin{figure*}[!ht]  \centering \small
    \includegraphics[width=0.98\linewidth,trim={40 50 90 50},clip]{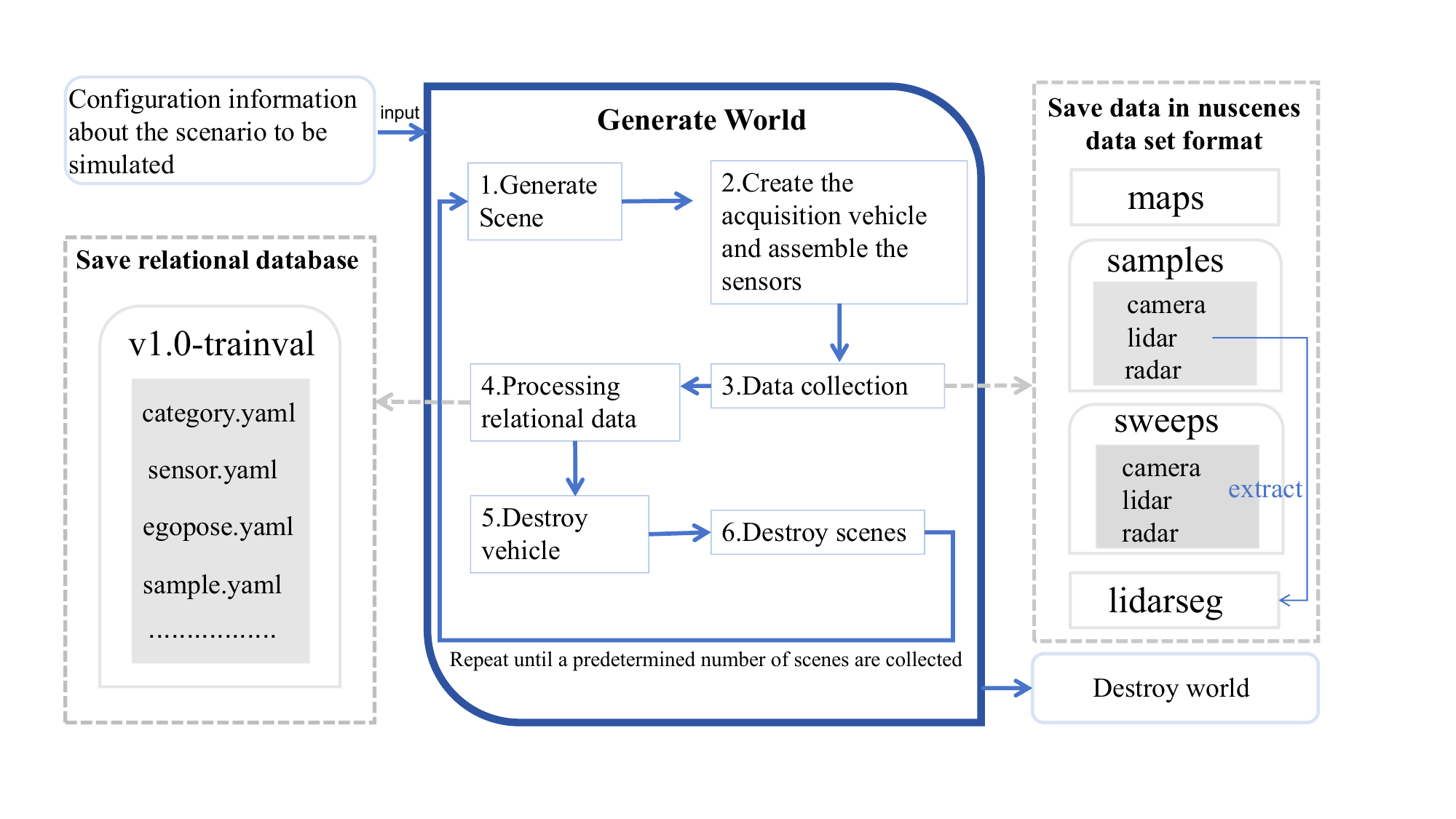}
    \caption{Data collection flow chart.
    We can obtain the map information needed for collection, the number of collection scenes, and vehicle configuration from the configuration file.
    Then, the procedure will generate a world to start collecting data.
    In this process, the procedure will save data in Nuscenes data set format and process relational data to save the relational database.
    In particular, the lidarseg file is extracted from the lidar file in the samples folder.}
    \label{fig:data collect}
\end{figure*}
\begin{figure*}[!ht]  \centering \small
     \includegraphics[width=0.98\linewidth,trim={40 100 10 50},clip]{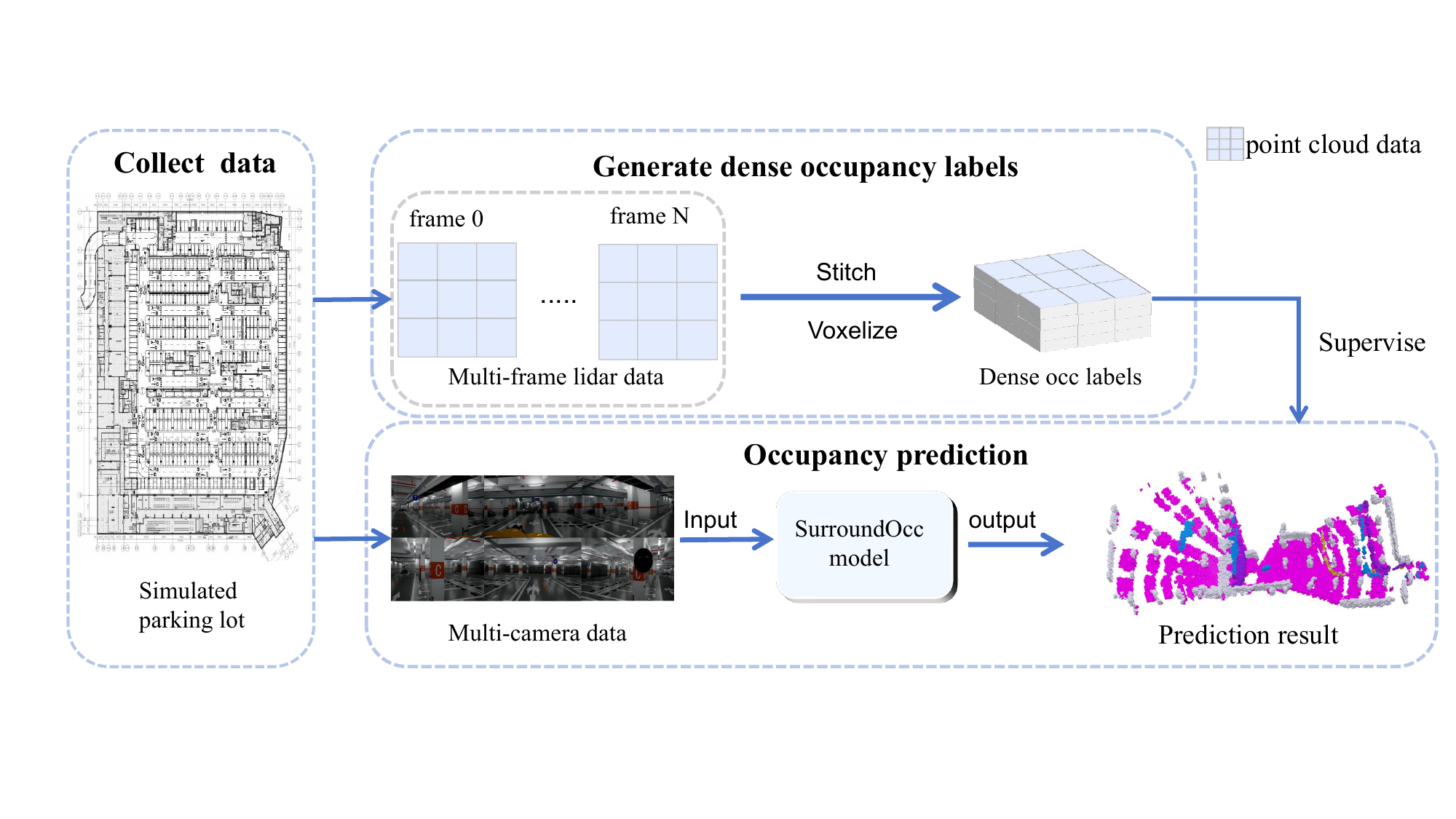}
    \caption{The pipeline of the proposed method.
    Initially, data collection is undertaken within a simulated parking lot environment.
    This process entails gathering point cloud data, which is subsequently utilized to produce detailed occupancy labels.
    Concomitantly, the camera footage encapsulated within the amassed dataset serves as the input for the SurroundOcc model, enabling it to perform occupancy prediction tasks, thereby harnessing the rich visual information for an enhanced understanding of the parking space.
    The generated dense occupancy labels supervise this process.}
    \label{fig:method}
\end{figure*}

This paper will introduce detailed methods for the above two problems. \autoref{fig:method} shows the process of the main methods.
\subsection{Data Collection}
Training SurroundOcc models requires data sets built with the nuScenes structure, upon that, the study configured the same sensors as nuScenes to ensure that the collected data is consistent with the nuScenes data set structure, which can also avoid the impact of training the network when the desired data is not found.
\autoref{fig:data collect} shows the flow chart of data collection.
 
\begin{figure}[!ht]  \centering \small 
    
    \includegraphics[width=1\linewidth]{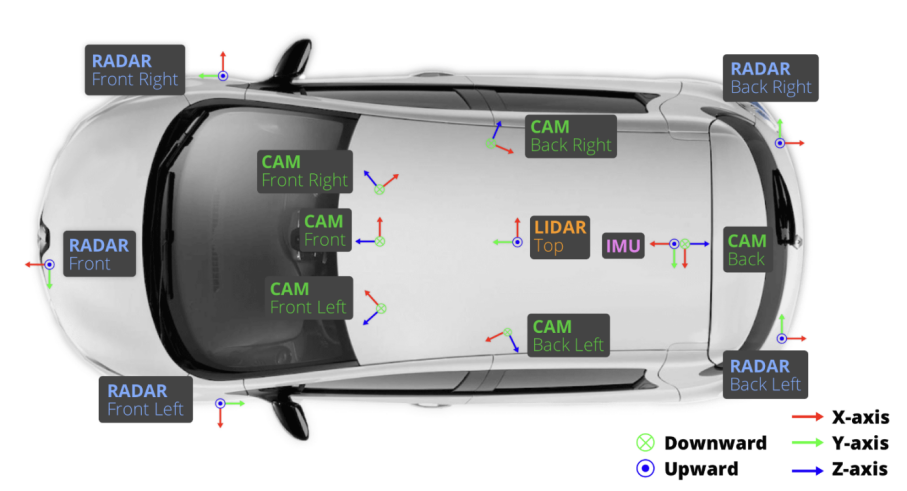}
    \caption{The sensor installation location of the nuScenes official acquisition vehicle \cite{caesar2020nuscenes}.
    Six cameras and five millimeter-wave radars are mounted in different directions to collect circumnavigation data.
    Lidar is mounted on the top of the car to scan the surrounding environment.}
    \label{fig:nuscenes vehicle}
\end{figure}

\begin{figure}[!ht]  \centering \small
    \includegraphics[width=1\linewidth]{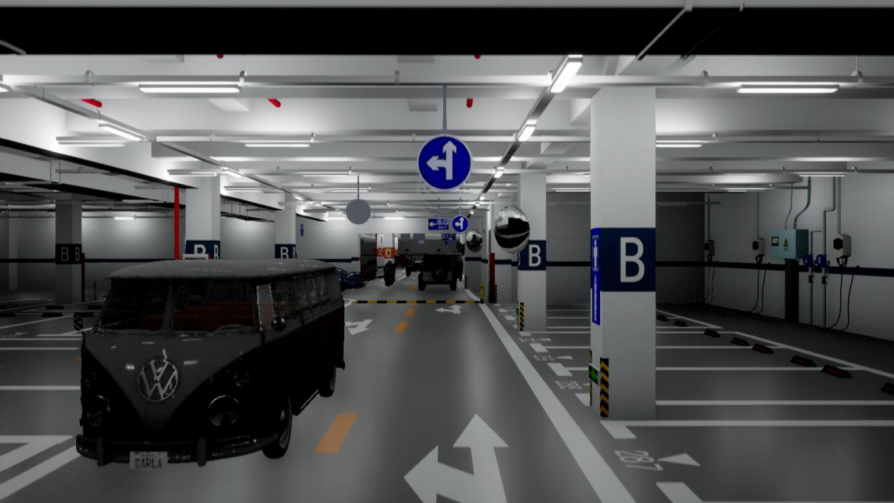}
    \caption{This is an image collected by camera.
    A frame of data collected during vehicle driving is displayed }
    \label{fig:camera image}
\end{figure}

Using the CARLA simulator, we can "install" several cameras, including lidar, IMU, millimeter-wave radar, and other simulation sensors on the vehicle.
We also set semantic lidar and other semantic sensors for GT data generation.
These simulation sensors have no physical structure and can be placed in any position of the vehicle. 
According to \autoref{fig:nuscenes vehicle} the sensor configuration of the acquisition vehicle given in the official document of nuScenes, the acquisition vehicle used in this research is equipped with six cameras, five millimeter-wave radars distributed in different directions, and one lidar on the top of the vehicle. 
In this study, the lidar was replaced by the semantic lidar provided by CARLA to facilitate the subsequent semantic tag extraction. \autoref{tab:camera},\autoref{tab:radar},\autoref{tab:lidar} shows camera parameters, millimeter wave radar parameters, and semantic Lidar parameters, respectively. 
\autoref{fig:camera image} shows a frame of data of the underground parking lot collected by the camera. 
This research employs a custom-built model of the underground parking facility at the College of Engineering, Southern University of Science and Technology, serving as the foundational parking lot map for the study.
\begin{table}[!ht]  \centering \small
  \caption{Six camera parameters for acquisition}
  \label{tab:camera}
  \resizebox{\columnwidth}{!}{
  \begin{tabular}{c|c|c|c|c}     \toprule
    camera name & location(x,y,z) & direction(yaw) & image resolution & fov \\ \midrule
    CAM\_FRONT & (1.5,0,2) & 0 & 1600x900 & 70 \\  \midrule
    CAM\_FRONT\_RIGHT & (1.5,0.7,2) & 55  & 1600x900 & 70 \\   \midrule
    CAM\_FRONT\_LEFT & (1.5,-0.7,2) & -55  & 1600x900  & 70\\    \midrule
    CAM\_BACK\_LEFT & (-0.7,0,2) & -110  & 1600x900 & 70 \\    \midrule
    CAM\_BACK & (-1.5,0,2) & 180  & 1600x900  & 110\\    \midrule
    CAM\_BACK\_RIGHT & (-0.7,0,2) & 110  & 1600x900 & 70 \\
    \bottomrule
  \end{tabular}
  }
\end{table}

\begin{table}[!ht] \small  \centering
  \caption{Five radar parameters for acquisition}
  \label{tab:radar}
  \resizebox{\columnwidth}{!}{
  \begin{tabular}{c|c|c|c|c}   \toprule
    radar name & loc(x,y,z) & dir(yaw) & horizontal\_fov & vertical\_fov \\  \midrule
    RADAR\_FRONT & (1.5,0,0.5) & 0 & 80 & 30 \\ \midrule
    RADAR\_FRONT\_RIGHT & (1.5,0.7,0.5) & 90  & 80 & 30 \\ \midrule
    RADAR\_FRONT\_LEFT & (1.5,-0.7,0.5) & -90  & 80 & 30 \\ \midrule
    RADAR\_BACK\_LEFT & (-1.5,-0.7,0.5) & 180  & 80 & 30 \\ \midrule
    RADAR\_BACK\_RIGHT & (-1.5,0.7,0.5) & 180  & 80 & 30 \\ \bottomrule
  \end{tabular}
  }
\end{table}

\begin{table}[!ht]  \centering \small
  \caption{Semantic Lidar parameters}
  \label{tab:lidar}
  \resizebox{\columnwidth}{!}{
  \begin{tabular}{c|c|c|c|c|c|c}   \toprule
    lidar name & loc(x,y,z) & dir(yaw) & number of channels
 & range & horizontal\_fov & vertical\_fov range \\  \midrule
    LIDAR\_TOP & (0,0,2) & 90 & 64 & 80m & 360° & (-30°, 10°) \\ \bottomrule
  \end{tabular}
   }
\end{table}

\subsubsection{Basic Configuration}
\begin{figure}[!ht]  \centering \small
    \includegraphics[width=1\linewidth]{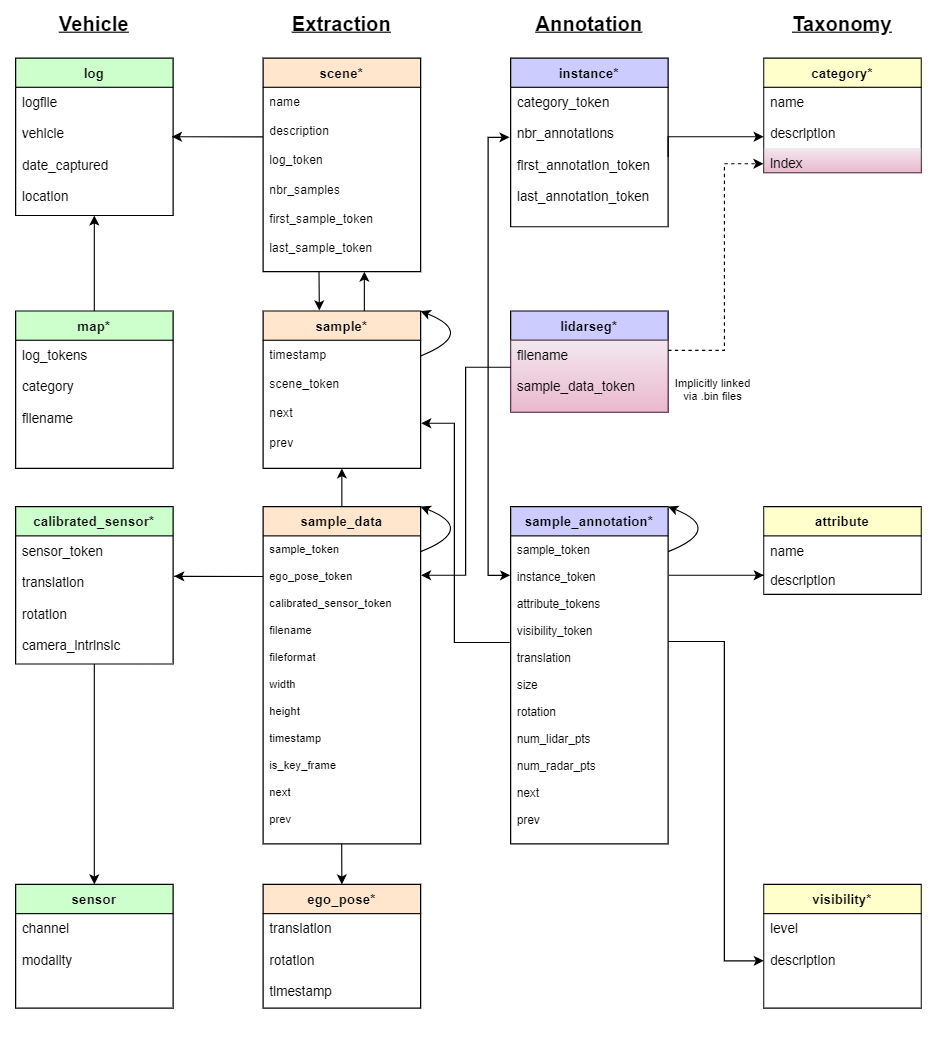}
    \caption{Relational database for the nuscenes dataset.
    The nuscenes dataset uses a relational database to manage the data, which contains 13 tables stored in JSON file format \cite{caesar2020nuscenes}.
    These tables include annotations and metadata, such as calibration, maps, vehicle coordinates.}
    \label{fig:relation}
\end{figure}

A centralized configuration approach is adopted to comprehensively configure all scenarios in the simulation, wherein all elements—including vehicles, sensors, and map details are consolidated within a single configuration file.
Specifically, the config.yaml file acts as an index, facilitating easy access to all other dispersed configuration information, thus ensuring streamlined management and adjustability of the entire simulation setup.
This file contains the information about the name of the folders to be created,
the IP address and port number to connect to the CARLA client, sensor configuration, map information, scene Settings, collection time settings, and the number of keyframes to be collected. 
NuScenes format folders were first created after reading the config file. 
Taking train data as an example, the acquired data folders include maps, samples, sweeps, v1.0-trainval and lidarseg. 
Maps stored the rasterized pictures of the map. 
The collected sensor data is archived into two distinct categories: Samples and Sweeps. 
Samples are repositories for keyframe data, encapsulating the most salient frames crucial for understanding the scene dynamics.
Conversely, Sweeps encompass a comprehensive record, retaining all sensor data captured over time.
Within the dataset’s structured framework, the v1.0-trainval directory houses numerous JSON files, each meticulously documenting metadata and annotations essential for training and validating the machine learning models.
Save data annotation information, vehicle self-pose information, map information, and sensor information refer to \autoref{fig:relation}. 
Each piece of information has a corresponding token for indexing to the corresponding data. 
Lidarseg holds semantic labels for point cloud data of key frames. 
\subsubsection{Scene Generation}
A scenario includes an autonomous driving situation, such as a vehicle turning, whether there are pedestrians on the side of the road. 
The process commences with the CARLA simulator parsing the configuration file to initiate scene generation.
Subsequently, it establishes a connection to the CARLA emulator utilizing the specified IP address and port number detailed within the configuration.
Upon successful connection, the bespoke map representing the underground parking lot of our case study is then seamlessly integrated into the virtual environment, thereby laying the groundwork for the intricate simulations to unfold \cite{lan2023virtual}.
Each scene is then created in the generated world. 
The scene information includes the location of the acquisition car, the weather conditions, the index of the sensor information installed on the car.
Within the scope of this study, a strategic approach has been adopted to ensure comprehensive coverage of the entire underground parking lot during data collection.
For this purpose, random points are strategically selected across different scenes, serving as spawn locations for autonomously placed vehicles, which we term "acquisition vehicles."
This intentional randomness in vehicle placement guarantees a diverse and representative sampling of the parking environment, thereby enriching the dataset and enhancing its fidelity to real-world conditions.
The location of vehicles can be generated in the map obtained through the API provided by CARLA. 

In each simulated scenario \cite{lan2019simulated}, a location is randomly chosen to spawn the acquisition vehicle, tasked with data collection. 
Additionally, other random positions are designated for generating a multitude of passing vehicles, thereby introducing a dynamic element and enhancing the overall diversity and realism of the simulated scenes.
This deliberate inclusion of varied traffic patterns ensures a thorough representation of potential interactions and complexities found within an actual underground parking lot, enriching the dataset for more robust analysis and model training.

Various sensors need to be installed on the acquisition vehicle after each scene is generated. 
In this case, cali-brated\_sensors.yaml can be indexed through config.yaml, which prestores the parameters of all sensors including camera position and rotation Angle, millimeter wave radar position and rotation Angle, semantic lidar channel number, acquisition frequency, and detection range. 
The control mode of the acquisition vehicle and all other vehicles is set to automatic driving after assembling the required sensor for the acquisition vehicle through this file. 
However, the vehicle may hit the wall after moving for a certain distance due to the imperfect automatic driving control provided by CARLA, thus reducing the efficiency of acquisition. 
Since every scene generation will refresh all vehicle
positions and start moving again, this study chooses to solve this problem by increasing the number of scenes and reducing the collection time of each scene, so as to ensure that most of the data is collected during the moving process of vehicles.

The acquisition process is carried out according to the frame count cycle, that is, each cycle represents a time frame.
The current frame count is printed at the beginning of each cycle during the collection process.
For every frame, if the number of frame count added one can be evenly divided by the result of the key frame interval divided by the fixed time interval, the current frame is considered to be a keyframe.
In the non-key frame, the collected data is updated directly. 
Under the keyframe, the sensor list is traversed to obtain the data of each sensor, and then the data list of the sensor is traversed further. 
For each data, the corresponding json file is updated. 
If it is the last data of the key frame, that is, semantic lidar data, a method will be called to extract the semantic label in the data and update it to the data set \cite{lan2022semantic}.
After processing a portion of the frame count data, each instance object in the collected data is traversed, a method is called to obtain its visibility, and the comment information for that instance is updated to the data set \cite{lan2018real,lan2019evolving}.
Finally, the data saved in all sensors is cleared for subsequent collection.

\subsubsection{Sensor Data Storage}
The data collected by the vehicle during the driving process is not stored at will, different files need to be processed separately.
Upon its initial execution, the data collection script automatically generates the necessary folder structure conforming to the nuScenes dataset standards.
The imagery captured by onboard cameras is systematically filed within the dedicated 'CAM' directory, whereas the millimeter-wave RADAR’s point cloud data is routed to a specifically established 'radar' folder.
Handling semantic LiDAR data necessitates additional processing steps.
Each point in the CARLA-supplied semantic LiDAR point cloud embodies six attributes: Cartesian coordinates (x, y, z), the cosine of the incidence angle, the index of the struck object, and the semantic label of that object.
This rich information is accessed by engaging with individual point objects within CARLA.

The process commences with identifying the point cloud’s channel count for archiving purposes.
Sequentially iterating through each channel, points are examined individually, with pertinent data extracted into a crafted point list.
These point lists are then aggregated into a comprehensive points aggregate, encapsulating all point data.
Utilizing NumPy's capabilities, this aggregated dataset is formatted into a multidimensional array and committed to disk in float32 format via the tofile method, each LiDAR dataset thereby becoming a .bin file akin to the nuScenes format.

In tandem with the point cloud, semantic labels demand separate extraction.
This involves parsing each point in the semantic LiDAR stream to isolate solely the semantic labels. 
These labels are then mapped to correspond with the nuScenes semantic tag, compiled into a list.
Table \autoref{tab:tag_mapping} outlines this meticulous label mapping schema. Employing NumPy once again, these semantic labels are serialized in uint8 format, ensuring each semantic label file complements its respective point cloud file as a .bin document.

Crucially, the dual application of NumPy's tofile method, each instance specifying distinct data types, guarantees impeccable alignment between the point cloud files and their corresponding semantic label files.
Consequently, every point within the point cloud can readily locate its matching semantic annotation, fostering a highly synchronized and interpretable dataset.
\begin{table}[!ht]  \centering \small
  \caption{Mapping from CARLA semantics to nuScenes semantics}
  \label{tab:tag_mapping}
  \resizebox{\columnwidth}{!}{
  \begin{tabular}{c|c|c|c}   \toprule
    CARLA semantic tag & nuScenes semantic tag & CARLA semantic & nuScenes semantic\\      \midrule
    0 & 0 & unlabeled & noise  \\ 
        \midrule
    1 & 24 & road & flat.driveable\_surface  \\ 
        \midrule
    2 & 26 & sidewalk & flat.sidewalk  \\ 
        \midrule
    3 & 28 & building & static.mamade  \\ 
        \midrule
    4 & 28 & wall & static.mamade  \\ 
        \midrule
    5 & 28 & fence & static.mamade   \\ 
        \midrule
    6 & 28 & pole & static.mamade   \\ 
        \midrule
    7 & 28 & traffic light & static.mamade   \\ 
        \midrule
    8 & 28 & traffic sign & static.mamade   \\ 
        \midrule
    9 & 30 & vegetation & static.vegetation  \\ 
        \midrule
    10 & 27 & terrain & flat.terrain  \\ 
        \midrule
    11 & 0 & sky & noise  \\ 
        \midrule
    12 & 2 & pedestrain & human.pedestrain.adult  \\ 
        \midrule
    13 & 14 & rider & vehicle.bicycle  \\ 
        \midrule
    14 & 17 & car & vehicle.car  \\ 
        \midrule
    15 & 23 & truck & vehicle.truck  \\ 
        \midrule
    16 & 16 & bus & vehicle.bus.rigid  \\ 
        \midrule
    17 & 15 & train &  vehicle.bus.rigid \\ 
        \midrule
    18 & 21 & motocycle & vehicle.motocycle  \\ 
        \midrule
    19 & 14 & bicycle & vehicle.bicycle \\ 
        \midrule
    20 & 29 & static & static.other  \\ 
        \midrule
    21 & 9 & dynamic & movable\_object.barrier  \\ 
        \midrule
    22 & 29 & other & static.other  \\ 
        \midrule
    23 & 29 & water & static.other  \\     
    \midrule
    24 & 24 & road line  & flat.driveable\_surface  \\     
    \midrule
    25 & 24 & ground &  flat.driveable\_surface   \\ 
        \midrule
    26 & 29 & brigde & static.other  \\     
    \midrule
    27 & 29 & rail & static.other   \\     
    \midrule
    28 & 29 & guard rail & static.other   \\     
    \midrule
    29 & 24 & parking lane & flat.driveable\_surface  \\    
    \midrule
    30 & 24 & parking area & flat.driveable\_surface  \\     
    \bottomrule
  \end{tabular}
  }
\end{table}
\subsubsection{Relational Database Storage}
While saving the collected data, it is also necessary to generate the corresponding relational database according to the data so that the required files can be accurately indexed while using the data set. 
The information in relational database is stored in multiple json files,
See \autoref{fig:relation}. 
In this research, the management of JSON files has been systematically categorized into three distinct processing streams to ensure strict adherence to the nuScenes dataset structure:
\begin{enumerate}[leftmargin=*]
    \item Save predefined data and sensor data information

    This type of json file holds predefined information such as maps, sensor configurations, scene information, and the storage address of the collected data.

    \item Save annotated information

    This type of file saves 3D surrounding box information for objects other than the car that appear in the collected data. This type of file saves 3D surrounding box information for objects other than the car that appear in the collected data.

    \item Saving vehicle and sensor pose information

    In this type of file, information including the position of the vehicle, the yaw Angle, and the position and yaw Angle of the sensor installed on the vehicle are saved in each frame during the driving of the vehicle.
\end{enumerate}

Before processing the three types of information, we need to design a method to generate the token of each piece of information.
Each json file contains multiple pieces of information, and each piece of information has a corresponding token for indexing other data to itself. 
Since tokens need to be generated separately, the generate\_token method is defined in this study to create tokens.
The method uses md5 hash algorithm to generate tokens.
Md5 is a commonly used hash algorithm that can convert data of any length into a fixed length hash value.
The method takes two parameters, key and data.
First, create a hashlib.md5 object using key as the initial value to compute the hash value.
Secondly, encode(' utf-8 ') converts key and data into UTF-8-encoded strings and passes them to an md5 object using the update() method to update the hash calculation.
Finally, the hexdigest method is called to obtain the calculated hash value and return it as a result. Take the sensor file as an example, the key and data used for the token generation are the 'sensor' string and the specific sensor name (e.g., camera, radar), respectively.

\paragraph{Save predefined data and sensor data information}

Pre-defined data refers to the information defined before the program runs, including semantic categories, sensor binding position on the vehicle, sensor parameters, and map information.
It can read the information from the files that save the information after the program starts running, and then generate separate tokens for it. 
Regarding sensor data handling, the process is streamlined to entail merely reading the storage addresses logged by each sensor. 

\paragraph{Save annotated information}
This process revolves around capturing the bounding box (in the global or world coordinate system) metadata for each object detected in keyframes.
It documents the object's position, dimensions, orientation, and classification \cite{lan2022class,gao2021neat} meticulously, forming a vital component of scene understanding for autonomous vehicles.
Determining the object's bounding box necessitates assessing its visibility from the perspective of the self-driving vehicle. 
The study introduces a custom method, get\_visibility, which leverages X-ray-like detection principles to gauge visibility levels based on the presence of observable points across varying directions.
The visibility score reflects the object's discernibility within the scene, with elevated scores indicating heightened observability.
The methodology unfolds as follows:
\begin{enumerate}[leftmargin=*]
    \item Sensor Identification

     It initiates by traversing the sensor list to pinpoint the 'semantic LiDAR' sensor, recognizing its pivotal role in detailed object perception.

     \item Target and Vehicle Positioning

     The system then retrieves the spatial coordinates of both the autonomous vehicle and the target instance for which visibility is to be assessed.

     \item Plane Construction and Ray Casting

     Focusing on the target's mid-Z plane, five strategic points are selected around the central point of the target's xy-plane projection.
     These points encompass the full breadth of the target's potential visibility profile.
     From the autonomous vehicle's vantage point, rays are virtually cast towards these five points, effectively probing visibility in five distinct directions.
\end{enumerate}
The method provides a nuanced understanding of how well each object is exposed to the sensing capabilities of the vehicle by quantifying visibility through this directional ray detection, thereby informing subsequent decision-making processes in simulation or real-world autonomous navigation.

For each ray, a set of ray points is obtained using the ray detection function.
The filter is then used to get points that are not in the bounding box between the car and the instance and are labeled NONE, that is, there is no other object between the car and the target. 
Calculate the number of visible points in each direction based on the number of filtered visible points.
Finally, the largest value of the number of visible points in each of the five directions is taken as the final number of visible points for instance, mapped to the visibility level, and the result is returned.
After obtaining the visibility level of the instance, it is converted into a string form to obtain the visibility identifier visibility\_token.
Next, get the size of the instance, assign its height, width, and length to the size list, convert xyz coordinates to whl coordinates, and get the 3D surround box information of the target.
All data is saved as a sample\_annotation.json file after processing.

\begin{figure}[!ht]  \centering \small
    \includegraphics[width=0.98\linewidth,trim={60 280 120 40},clip]{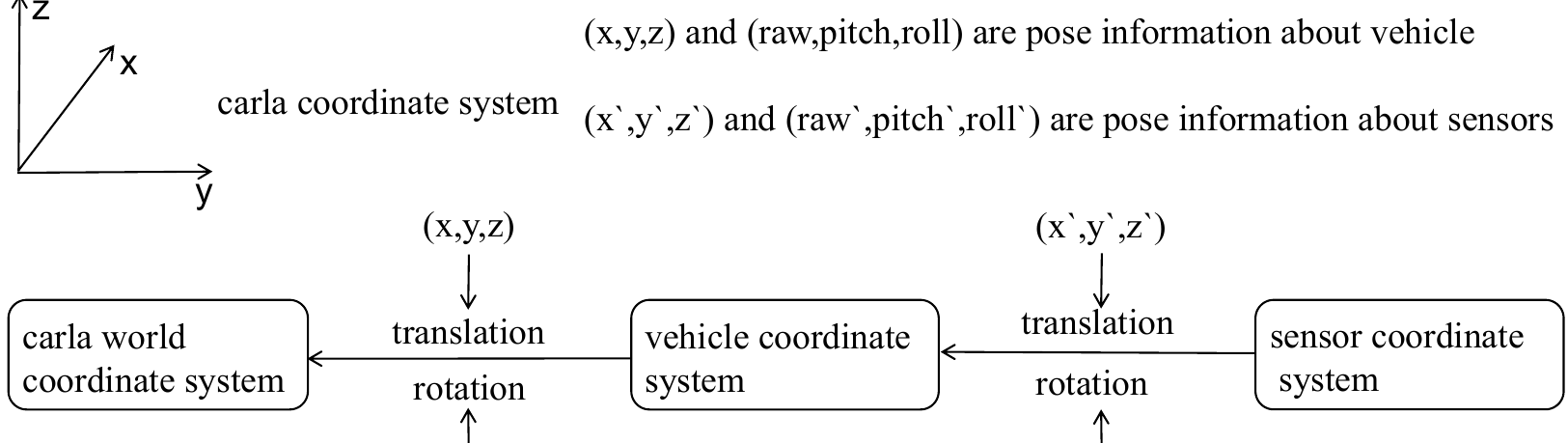}
    \caption{Diagram of the conversion process between coordinate systems in CARLA.
    Using translation and rotation, we can convert the sensor coordinate system to the car coordinate system and the car coordinate system to the world coordinate system.
    The sensor coordinate system cannot be directly converted to the world coordinate system}
    \label{fig:carla translate}
\end{figure}

Algorithm \autoref{IVC} is the algorithm pseudocode to get the instance visibility:
\begin{algorithm}
\caption{Instance visibility calculation}
\label{IVC}
\begin{algorithmic}[1]
\Procedure{input}{Instance object}
\State $visible\_count \gets 0$ \Comment{Count starts at 0}
\For{$i \gets 1$ to $5$}
\State  A ray is cast from the agent to a target point, undergoing 5 transformations based on instance positions.
\State Points in ray points are filtered if outside the autonomous vehicle's bounding box, the target instance's box, or labeled as NONE.
  \If{$points$ is None}
    \State $visible\_count+1$ 
  \EndIf
\EndFor
\State \Return dict\{visible\_count\} 
\EndProcedure
\end{algorithmic}
\end{algorithm}
\paragraph{Save vehicle and sensor position information}
The pose information of the vehicle contains the rotation matrix and the displacement information that converts the vehicle coordinate system into the world coordinate system under this frame. 
The sensor's position and pose information contains the rotation matrix and displacement information from the sensor coordinate system to the vehicle coordinate system.
The use of this information is shown in \autoref{fig:pcd stitch}.
In this paper, a scheme is designed to calculate pose information.
Since the data in this study is collected in the CARLA simulation world, the relevant pose information needs to be obtained in CARLA's multiple coordinate systems.
In the CARLA simulation world, the world coordinate system, the vehicle coordinate system and the coordinate system of various sensors have the same direction, and they are all left-handed systems.
Converting between coordinate systems essentially requires only accounting for the translation amounts and the rotation angle around the Z-axis.
In this study, the translation required for conversion between coordinate systems is set as $T(translation)$,the rotation matrix is set as $R(rotation)$.
Take the conversion of the vehicle coordinate system to the world coordinate system as an example. 
Firstly, CARLA API can be used to obtain the vehicle's position$(x,y,z)$, yaw Angle, slope pitch and rollover degree roll in the world coordinate system, then the translation vector $T$ can be represented as $[x,y,z]$, and the rotation matrix can be represented as $R=[R_3(roll)R_2(pitch)R_1(yaw)]$, Where $R_3,R_2,R_1$ represent the rotation matrix calculated for roll, pitch and yaw respectively, and the final rotation matrix can be obtained by multiplying them left from right to left.
The specific process is shown in \autoref{fig:carla translate}, and the position and pose information of the sensor can be obtained similarly. 
After obtaining the pose information of the two, they can be allocated independent tokens to save.

\subsection{SurroundOcc Method Design}
\subsubsection{Generation of Dense Occupancy Ground Truth}
In the training process, the direct employment of sparse LiDAR point clouds as supervisory signals proves inadequate for predicting sufficiently dense occupancy maps.
Consequently, SurroundOcc devises a methodology that generates dense occupancy information by leveraging a multi-frame point cloud aggregation.
The procedure initiates with the concatenation of point cloud datasets from several static scene frames and dynamic object instances. 
Subsequently, Poisson surface reconstruction is employed to interpolate the inter-point gaps, thereby yielding a high-density point cloud dataset.
Following this, both the reconstructed and initial sparse point clouds undergo voxelization.
The nearest neighbor algorithm is employed to assign semantic labels to the now-dense voxel grid, which transfers labels from the voxelized sparse point cloud to its more granular counterpart, ensuring a densely annotated occupancy map.

\begin{figure}[!ht]  \centering \small
    \includegraphics[width=0.98\linewidth,trim={10 0 0 10},clip]{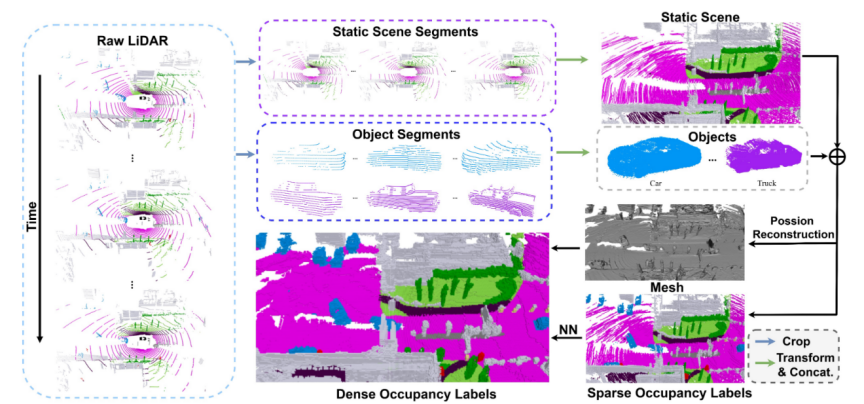}
    \caption{SurroundOcc generates dense occupancy truth process diagrams \cite{wei2023surroundocc}. 
    Firstly, It fuses dynamic object and static scene LiDAR scans. 
    The method then employs Poisson reconstruction to fill holes and voxelizes the mesh to create dense occupancy labels.}
    \label{fig:value generation}
\end{figure}

In our work, this methodology is replicated specifically to synthesize comprehensive occupancy ground truths for underground parking scenarios, serving as a pivotal component in the training regimen. 
\autoref{fig:value generation} illustrates the schematic overview of this label generation technique.
\paragraph{Multi-frame Point Cloud Stitching}
Within this framework, the initial step involves consolidating multi-frame point clouds. 
This consolidation is essential as it harnesses diverse viewpoints, thereby maximizing the capture of scene information and ensuring a comprehensive representation of the environment.
The most intuitive approach is to convert the multi-frame radar point sequence in a scene directly into a unified coordinate system, and then connect it.
However, such a method only applies to completely static scenes and ignores moving objects, so static scenes and dynamic objects need to be dealt with separately.

\begin{figure}[!ht]  \centering \small
    \includegraphics[width=0.98\linewidth,trim={15 60 260 120},clip]{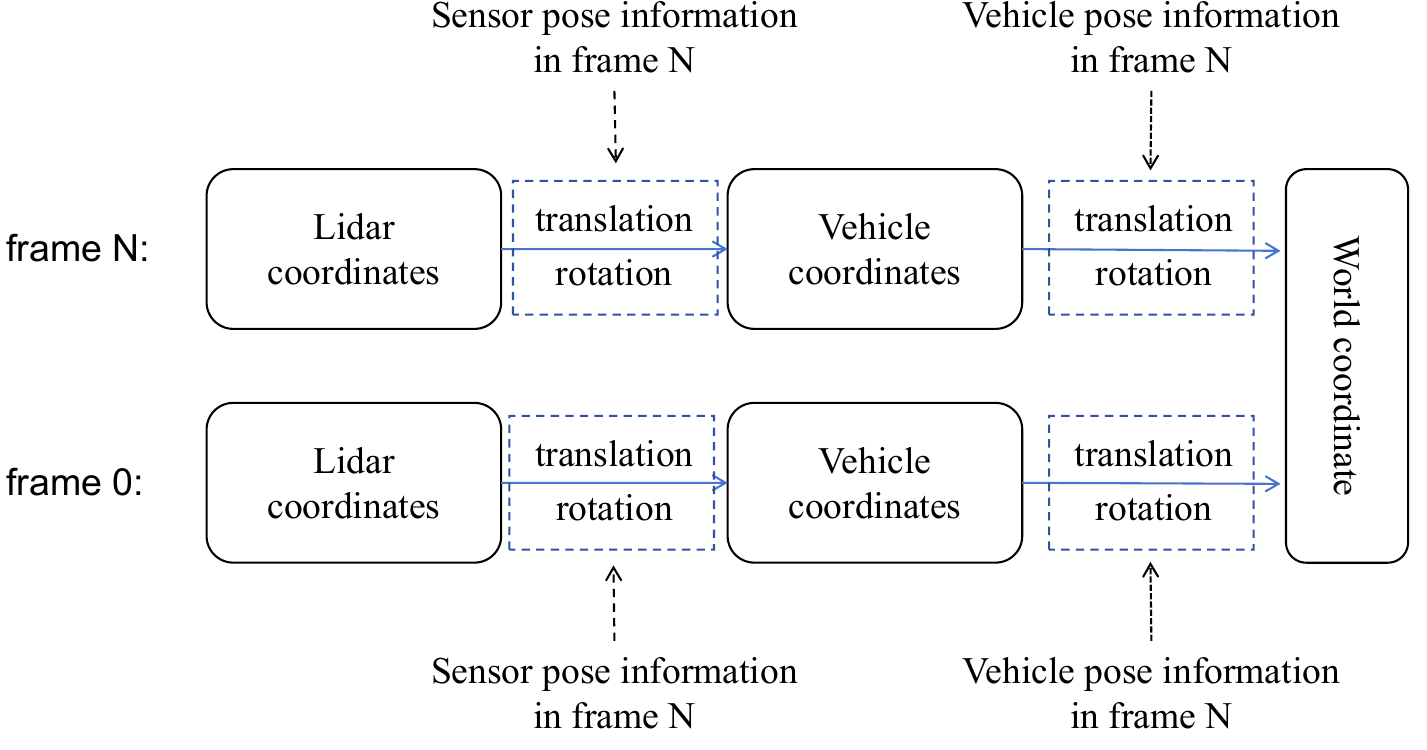}
    \caption{This figure shows how to convert the frame N point cloud data coordinates to frame 0.
    The point cloud coordinates can be converted to the world coordinates by using the sensor pose information and self-parking pose information of the Nth frame.
    These two pieces of information are also used to convert the coordinate information from the world coordinates to the point cloud coordinates of frame 0
    }
    \label{fig:pcd stitch}
\end{figure}

Firstly, for each frame of point cloud data in a sequence, the point cloud of movable objects is separated from the point cloud according to the 3D boundary frame label so as to obtain the point cloud of static field spot and dynamic object in the multi-frame data, respectively.
The collected static scene segmentation and dynamic object segmentation are integrated into two sets respectively after processing all the point cloud data in a sequence. 
Upon acquiring the multi-frame point cloud dataset, which has been processed in sequence, the subsequent action entails aligning these frames within a unified spatial context. 
Each frame's coordinate system must be transformed into the global or world coordinate system to achieve this integration.
This critical transformation is facilitated through the application of calibration matrices and the determination of each sensor's self-pose, thereby ensuring coherent fusion of the point cloud data across all frames.
Represents the full 3D point cloud of the current frame as $\mathcal{P}$.
Then, the converted static scene segmentation is expressed as $\mathcal{P}_{ss}=\{\mathcal{P}^1_{ss},\mathcal{P}^2_{ss},...\mathcal{P}^n_{ss}\}$.
Dynamic object segmentation can be expressed as $\mathcal{P}_{os}=\{\mathcal{P}^1_{os},\mathcal{P}^2_{os},...\mathcal{P}^m_{os}\}$.
Where n and m are the number of frames and objects in the sequence, respectively.At this point, the entire static scene can be represented as $\mathcal{P}_s=[\mathcal{P}^1_{ss},\mathcal{P}^2_{ss},...\mathcal{P}^n_{ss}]$.
Dynamic objects are represented as $\mathcal{P}_o=[\mathcal{P}^1_{os},\mathcal{P}^2_{os},...\mathcal{P}^m_{os}]$.
Finally, according to the position of the dynamic object in the current frame and the pose of the car, the static scene and dynamic object can be fused to obtain the 3D point cloud of the current frame: $\mathcal{P}=[\mathcal{T}_s(\mathcal{P}_s),\mathcal{T}_o(\mathcal{P}_o)]$,$\mathcal{T}_s$ and $\mathcal{T}_o$ represent the transformation of static scenes and dynamic objects from the world coordinate system to the frame 0 point cloud coordinate system, respectively.
The whole process of coordinate conversion can be seen in \autoref{fig:pcd stitch}. In this way, the splicing of multi-frame point clouds can be completed, and then the semantic label of the key frame point cloud can be applied to the point cloud sequence in the subsequent work.
\paragraph{Densifying with Poisson Reconstruction}
Despite the integration of multiple frame point clouds, the resultant point cloud 
$\mathcal{P}$ continues to exhibit insufficient density.
Gaps persist between points, and their distribution remains uneven, a consequence of the inherent constraints of the LiDAR's beam configuration.
To address this, SurroundOcc proposes using Poisson reconstruction to fill the gaps between point clouds to get denser point cloud data.
Poisson reconstruction is an implicit surface reconstruction scheme in which the input is a set of directed point clouds and the output is a three-dimensional mesh. Compared with direct grid reconstruction such as Delaunay, the main difference is that the vertices of the output grid do not need to come from the original point cloud, the result is smoother, and the water tightness of the grid can be guaranteed due to global solution.
In order to achieve the goal, the first step is to calculate the normal vector according to the spatial distribution of the local neighborhood, and then the point cloud $\mathcal{P}$ is reconstructed into a triangular mesh $\mathcal{M}$ through the Poisson surface reconstruction method. 
The input of the algorithm is the point cloud data with normal vectors, and the output is a triangular mesh.
The resulting triangular mesh can obtain dense point clouds by filling the holes between point clouds with evenly distributed vertices, and then the processed point clouds can be voxelized to get $\mathcal{V}_d$.

\paragraph{Semantic Labeling with NN Algorithm}
Although intensive voxel occupancy is obtained through a series of methods, each voxel does not contain semantic labels, which is due to the fact that Poisson reconstruction only applies to 3D space and not semantic space. 
The nearest neighbor search algorithm can solve this problem well.

First, the point cloud $\mathcal{P}$ with semantics is voxelized to get $\mathcal{V}_s$, which will be more sparse than $\mathcal{V}_d$ due to the limited liDAR beam.
For each voxel in $\mathcal{V}_d$, the nearest neighbor algorithm is used to search for the voxel closest to $\mathcal{V}_s$ and assign its semantic label to itself,
so that all occupying voxels in $\mathcal{V}_d$ can obtain their own semantic label from $\mathcal{V}_s$. 
Specifically, the algorithm inputs all vertices of $\mathcal{V}_s$ and $\mathcal{V}_d$, 
first creates an empty PointCloud object pcd, assigns the vertices of $\mathcal{V}_d$ to pcd.points, 
and treats the vertices in $\mathcal{V}_d$ as the point cloud data of pcd. 
A kdtree object is then created using the KDTreeFlann algorithm to quickly search for nearest neighbor vertices.
Go through each vertex in $\mathcal{V}_s$, using the kdtree.search\_knn\_vector\_3d function to find the vertex closest to the current vertex in $\mathcal{V}_s$,
return the index of the closest vertex,
add the distances to the indices and distances lists, respectively, 
return these two lists. 
After obtaining the semantics of all vertices in $\mathcal{V}_d$ and the corresponding index of $\mathcal{V}_s$,
semantic information can be extracted from $\mathcal{V}_s$,
and finally connected with $\mathcal{V}_d$ to get a new voxel data,
in which semantic information is added to $\mathcal{V}_d$ as a new column.

Algorithm \autoref{NN algotithm} is the pseudocode for the nearest neighbor algorithm

\begin{algorithm}[!ht] \centering \small
\caption{Nearest Neighbor Algorithm}
\label{NN algotithm}
\begin{algorithmic}[1]
\Procedure{input}{verts1,verts2}
\State $indices \gets empty\_list$
\State $distances \gets empty\_list$ 
\State $pcd \gets verts1.location $ 

\Comment{Create an empty KD tree object kdtree; Initialize the kdtree using the point cloud data pcd}
\For{$vert \gets i$ to $verts2$}
\State  $ $inds, dist =kdtree.search(vert,1) 
\State append inds[0] to indices 
\State append dist[0] to distances 
\EndFor
\State \Return indices\  distances 
\EndProcedure
\end{algorithmic}
\end{algorithm}

\subsubsection{Overview of the SurroundOcc Model}
For the occupation prediction of the scene, this study needs to use multiple camera images as input to predict the 3D occupation of the scene, so this task can be expressed as: 
{\small 
\begin{equation} \centering
    \mathcal{V}=\mathcal{G}(\mathcal{I}^1,\mathcal{I}^2,...,\mathcal{I}^\mathcal{N})   
    \label{eq:xxx}
\end{equation}
}                                                           

\begin{figure}[!ht]  \centering \small
    \includegraphics[width=0.98\linewidth,trim={10 0 10 10},clip]{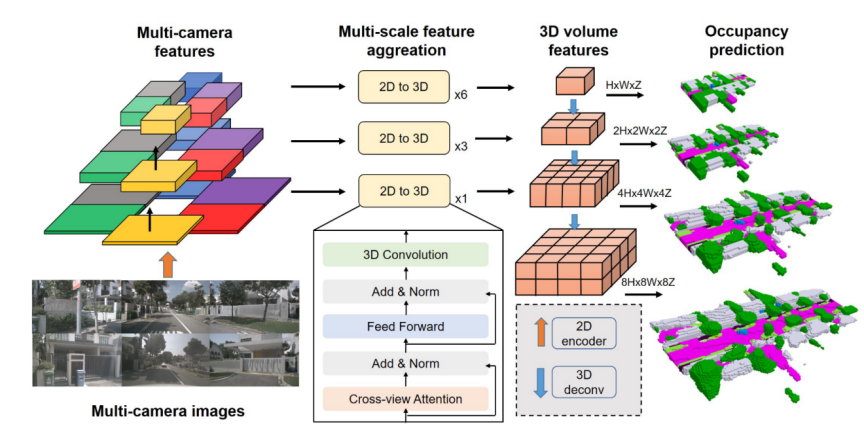}
    \caption{SurroundOcc model structure. First, it extracts multi-scale features of multi-camera images \cite{wei2023surroundocc}.
    Then, the model fuses multi-camera information and constructs 3D volume features in a multi-scale fashion.
    Finally, the 3D deconvolution layer is used to upsample 3D volumes and occupancy prediction is supervised in each level.}
    \label{fig:model structure}
\end{figure}

{\small $\mathcal{G}$} is a neural network,$\mathcal{I}^1$and$\mathcal{I}^\mathcal{N}$ represent the N images for inputs. 
The 3D occupancy is expressed as $\mathcal{V} \in \mathcal{R}^{\mathcal{H}\times \mathcal{W}\times \mathcal{Z}}$.
The value of $\mathcal{V}$ is between 0 and 1, indicating the occupancy probability of the grid.
Raising $\mathcal{V}$ to an $(\mathcal{L},\mathcal{H},\mathcal{W},\mathcal{Z})$ tensor yields the 3D semantic occupancy, where $\mathcal{L}$ represents the number of classes, and a class number of 0 indicates that the network has no occupancy.
\autoref{fig:model structure} shows the structure of the model. 
The model first receives a set of multi-camera images of the surrounding environment, and then uses the backbone network ResNet-101\cite{He_Zhang_Ren_Sun_2016} to extract multi-scale features of $\mathcal{N}$ cameras and $\mathcal{M}$ levels:$\mathcal{X}=\{\{\mathcal{X}_i^j\}^\mathcal{N}_{i=1}\}^\mathcal{M}_{j=1}$.
For each level, multiple camera features and spatial cross-attention are fused, and the output of the spatial cross-attention layer is a 3D voxel feature rather than a BEV feature.
Next, 3D convolutional networks are used to up-sample and fuse multi-scale voxel features. 
The occupancy prediction for each layer is overseen by the generated dense occupancy truth value and decaying loss weight.
\paragraph{2D-3D Spatial Attention}
Many 3D scene reconstruction methods integrate multi-view 2D features into 3D spaces by reprojecting 2D features back into 3D voxels using known poses. 
The calculation method of its voxel features is simply to average all the 2D features in the voxel. 
However, this method assumes that the contribution of different views to the 3D voxel is the same, and does not take into account the occlusion or blurriness in some views.
SurroundOcc uses spatial cross-attention to address this, specifically a cross-attention mechanism to fuse multiple camera features.
Firstly, the 3D reference points are projected into the 2D view and then the deformable attention mechanism \cite{Zhu_Su_Lu_Li_Wang_Dai_2020} is used to query the points and aggregate the information. 
In the query, a 3D voxel query is used instead of a 2DBEV query, which can better retain the 3D spatial information. 
A 3D voxel query can be defined as $\mathcal{Q} \in \mathcal{R}^{\mathcal{C} \times \mathcal{H} \times \mathcal{W} \times \mathcal{Z}}$. 
For each query, the model uses only the view in the 3D reference click, projects the corresponding 3D points into the 2D view according to the given internal and external parameters, and then samples 2D features around these projected 2D positions.
The output $\mathcal{F} \in \mathcal{R}^{\mathcal{C} \times \mathcal{H} \times \mathcal{W} \times \mathcal{Z}}$ of the cross attention layer is the weighted sum of the sampling features calculated according to the deformable attention mechanism:
\begin{equation} \centering \small
    DeformAttn(q, p, x)=\sum_{i=1}^{N_{head}}\mathcal{W}_i\sum_{j=1}^{N_{key}}\mathcal{A}_{ij}\cdot \mathcal{W}_i^{'} x(p+\Delta p_{ij})
    \label{eq:xxx}
\end{equation}

\begin{equation} \centering \small
     \mathcal{F}^p=\frac{1}{\left | \mathcal{V}_{hit} \right |}\sum_{i \in V_{hit}}DeformAttn(\mathcal{Q}^p,\mathcal{P}(q^p,i),\mathcal{X}_i)
     \label{eq:xxxx}
\end{equation}

$\mathcal{F}^p$ and $\mathcal{Q}^p$ refer to the $p_{th}$ element in the output feature and the 3D voxel query, respectively.
$q^p$ corresponds to the 3D position of the query,$\mathcal{P}$ represents the method of 3D projection to 2D.
$\mathcal{V}_{hit}$ is the view hit by the 3D query point,  $\mathcal{W}_i$ and $\mathcal{W}_i^{'}$ are the learnable weights. 
$\mathcal{A}_{ij}\in [0,1]$ is the attention weight calculated by the dot product of the query and key.
$ x(p+\Delta p_{ij})$ is a 2D feature in position $p+\Delta p_{ij}$.
Instead of performing expensive 3D self-attention, the model uses 3D convolution to interact features between adjacent 3D voxels.

\paragraph{Multi-scale Occupancy Prediction}
The model further extends spatial cross-attention to a multi-scale approach. 
Since 3D scene reconstruction requires more low-level features to help the network learn fine-grained details, a 2D-3D U-Net architecture is designed in the model.
Specifically, for a given 2D feature $\{\{\mathcal{X}_i^j\}_{i=1}^\mathcal{N}\}_{j=1}^\mathcal{M}$, multi-scale 3D voxel feature $\{ \mathcal{F}_j \in \mathcal{R}^{\mathcal{C}_j \times \mathcal{H}_j \times \mathcal{W}_j \times \mathcal{Z}_j}\}_{j=1}^\mathcal{M}$ is extracted using a different number of spatial cross-attention layers.
Then the 3D deconvolution layer is used to upsample the 3D voxel feature $ \mathcal{Y}_{j-1}$ of layer $j-1$ and fuse it with $\mathcal{F}_j$:
\begin{equation} \centering \small
       \mathcal{Y}_j=\mathcal{F}_j+Deconv(\mathcal{Y}_{j-1})
    \label{eq:xxx}
\end{equation}
For each layer, the network outputs an occupancy prediction with a different resolution of $\mathcal{V}_j \in \mathcal{R}^{\mathcal{C}_j \times \mathcal{H}_j \times \mathcal{W}_j \times \mathcal{Z}_j}$.
In order to obtain powerful high and low level 3D features, the network is supervised at every scale.
The model uses the cross-entropy loss and scene class affinity loss introduced by MonoScene\cite{cao2022monoscene} as supervisory signals.
For 3D semantic occupancy prediction, the model adopts multi-class cross-entropy loss, for the 3D scene reconstruction task, it is changed to two types of cross-entropy loss. 
Since high-resolution occupancy prediction is more important, the model uses the loss weight of decay $\alpha_j=\frac{1}{2^j}$ to monitor the $j_{th}$ layer.

%%%%%%%%%%%%%%%%%%%%%%%%%%%%%%%%%%%%%%%%%%%%%%
\section{Experiments}
\label{sec:experiments}
%%%%%%%%%%%%%%%%%%%%%%%%%%%%%%%%%%%%%%%%%%%%%%
\subsection{Experimental Environment Settings}
\subsubsection{Hardware Configuration}
Current Occupancy networks training is typically computationally intensive and is typically completed on GPU clusters.
The SurroundOcc models used in this study also require a large amount of computing resources for training, the occupancy models used in underground parking lots are trained in the Computer Department of Southern University of Science and Technology.
\autoref{tab:hardware} lists the hardware parameters of the GPU cluster.
\begin{table}[!ht]  \centering \small
  \caption{hardware configuration}
  \label{tab:hardware}
  \begin{tabular}{c|c}    \toprule
    hardware type & specification
 \\  \midrule
    GPU & NVIDIA L40s x2  \\ 
    RAM & 48GB x2 \\   \bottomrule
  \end{tabular}
\end{table}
\subsubsection{Model Hyperparameter}
\autoref{tab:Model hyperparameter} shows the hyperparameter Settings for the model. In this paper, two NVIDIA L40s graphics cards were used for training, with a total of 36 epochs.
The initial learning rate was 0.0002 and the learning rate was 0.1 in the backbone part. 
There are 60 training scenarios, 20 verification scenarios, and a total of 80 training data scenarios.
\begin{table}[!ht]  \centering \small
\setlength\tabcolsep{1pt} \renewcommand{\arraystretch}{1.0}
  \caption{hardware configuration}
  \label{tab:Model hyperparameter}
  \begin{tabular}{c|c}   \toprule
    type & parameter \\   \midrule
    optimizer & AdamW  \\  \midrule
    learning rate & 2e-4 \\  \midrule
    warmup\_iters & 500 \\  \midrule
    warmup\_ratio & 1.0/3\\   \midrule
    total\_epoches & 36\\  \bottomrule
  \end{tabular}
\end{table}
\subsection{Evaluation Metrics}
For 3D semantic occupancy prediction of the underground parking lot, IoU and mIoU are used as evaluation indexes in this paper.
For the occupation prediction of the scene, 
the semantic label of voxel is ignored, and the intersection ratio of the predicted sample and the actual sample (IoU) is used as the evaluation index.
The IoU metric measures the degree of overlap between the predicted segmentation results and the actual segmentation results.
It can be calculated by the following formula: IoU=(intersection area of the predicted result and the true result)/(union area of the predicted result and the true result).
All semantic categories are considered for the semantic occupation prediction of the scene, and mIoU is used as the evaluation index. 
The mIoU metric is an average calculation of the IoU of multiple samples and is used to evaluate the performance of the entire dataset. 
The steps to calculate mIoU are as follows:
\begin{enumerate}[leftmargin=*]
    \item Calculate the IoU for each sample in the data set.
    \item Add the IoU for all samples.
    \item Divide the sum by the number of samples to get the average IoU.
\end{enumerate}
In SurroundOcc, $\mathcal{TP}$, $\mathcal{FP}$ and $\mathcal{FN}$ represent TruePositive (true), FalsePositive (false) and FalseNegative (false negative), respectively.
They are indicators used to evaluate the accuracy and completeness of model prediction results in object detection or segmentation tasks.

$\mathcal{TP}$ (TruePositive) : refers to the number of samples that the model correctly predicts to be positive. 
That is, the number of positive cases that the model judges as positive cases.
$\mathcal{FP}$ (FalsePositive) : refers to the number of samples that a model incorrectly predicts as a positive example.
That is, the number of negative cases judged by the model as positive cases.
$\mathcal{FN}$ (FalseNegative) : refers to the number of samples in which the model incorrectly predicts negative cases. That is, the number of positive cases judged by the model as negative cases.

These indicators can be calculated by comparing them with real labels.
Suppose there are N samples, where the intersection of the prediction result and the real label is $\mathcal{TP}$,
the part of the prediction result that is a positive example but empty set with the real label is $\mathcal{FP}$, 
and the part of the real label that is a positive example but empty set with the prediction result is $\mathcal{FN}$, 
then the formula for calculating these indicators is as follows:
$\mathcal{TP}$= Intersection of the predicted result with the real label

$\mathcal{FP}$= the part of the predicted result that is a positive example but the union with the true label is an empty set

$\mathcal{FN}$= Evaluation index of the final model of the part of the real label that is a positive example but the union with the predicted result is an empty set

IoU and mIoU are calculated as follows:

\begin{equation} \centering \small 
    IoU=\frac{\mathcal{TP}}{\mathcal{TP}+\mathcal{FP}+\mathcal{FN}}
    \label{eq:xxx}
\end{equation}
     
\begin{equation} \centering \small
    mIoU=\frac{1}{\mathcal{C}}\sum_{i=1}^\mathcal{C}\frac{\mathcal{TP}_i}{\mathcal{TP}_i+\mathcal{FP}_i+\mathcal{FN}_i}
    \label{eq:xxx}
\end{equation}

\subsection{Experimental Procedure}
In order to verify the prediction effect of the model in the parking lot scene, relevant experiments were carried out in this study:
\begin{enumerate}[leftmargin=*]
    \item SurroundOcc pretraining models were first used to predict the ineffectiveness of underground parking lots.
    \item SurroundOcc models are trained using collected underground parking data to validate the effectiveness of the proposed approach.
\end{enumerate}

In this paper, the self-made nuScenes data set of parking lot scenarios is used. There are 80 scenarios in the training data,
of which 75\% are training sets and 25\% are verification sets. The training set contains 60 scenarios with 1200 keyframes.
The validation set contains 20 scenarios with 400 keyframes.

A comprehensive set of densely occupied voxel representations for underground parking lots was acquired by employing SurroundOcc to produce dense occupancy ground truth values, serving as the training dataset for further enhancing the model's capabilities.

\begin{figure}[!ht]  \centering \small
    \includegraphics[width=0.98\linewidth,trim={40 60 50 50},clip]{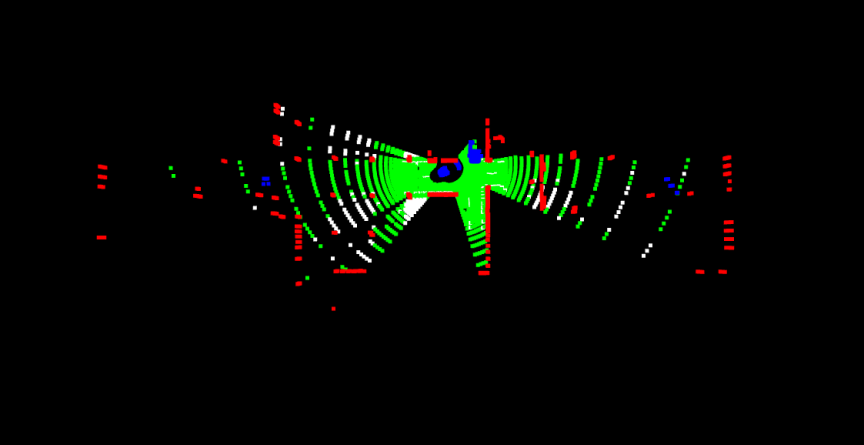}
    \caption{Visualization of point cloud files.
    The red part is the pillar and mammade.
    The green part is the groud and vehicles are shown in blue.}
    \label{fig:visable pcd}
\end{figure}

\begin{figure}[!ht]  \centering \small
    \includegraphics[width=0.98\linewidth,trim={90 20 120 80},clip]{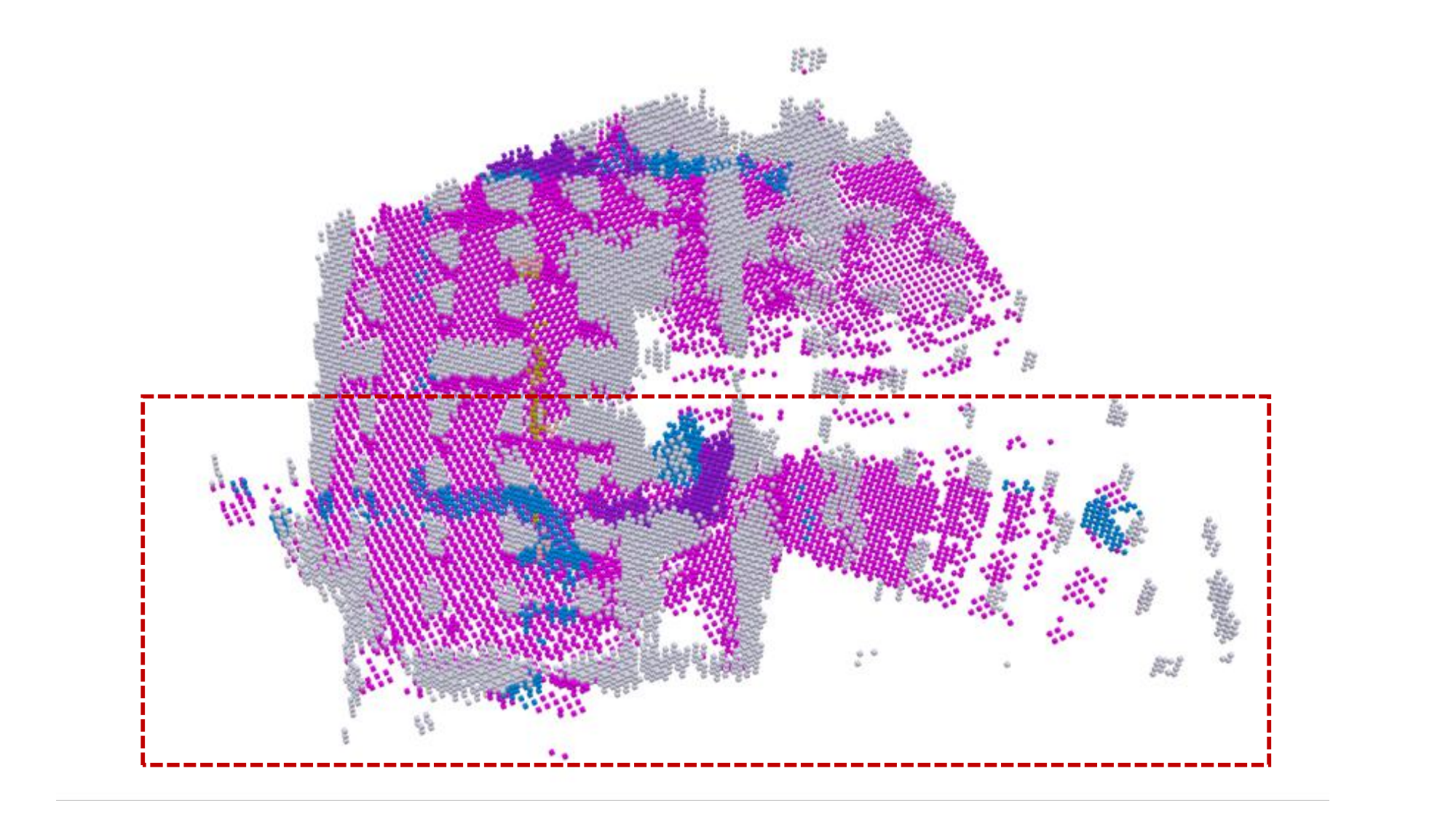}
    \caption{The occupancy truth value generated after completing multi-frame point cloud data. 
    In the visualization, the ground is denoted by the red region, while white signifies columns and walls.
    Blue and purple are used to depict vehicles. The area enclosed within the red box represents the original point cloud data, with the complementary segment stemming from the integration of additional frames' point clouds within the scene.
    }
    \label{fig:occ truth value}
\end{figure}

\begin{figure}[!ht]  \centering \small
    \includegraphics[width=0.98\linewidth,trim={0 0 0 10},clip]{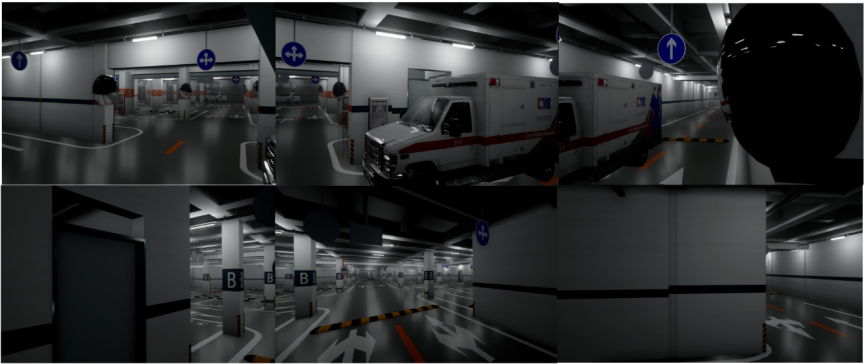}
    \caption{Multi camera pictures corresponding to the point cloud.
    Corresponds to the point cloud data.
    The order from left to right and top to bottom is: front left, front, front right, back left, back, back right}
    \label{fig:multi camera}
\end{figure}

\autoref{fig:visable pcd},\autoref{fig:occ truth value},\autoref{fig:multi camera} respectively show the LiDAR scanning point cloud in a certain frame, the corresponding occupation truth value generated, and the corresponding camera picture (picture sequence is front left, front, front right, back left, back right).

\begin{figure}[!ht]  \centering \small
    \includegraphics[width=0.98\linewidth]{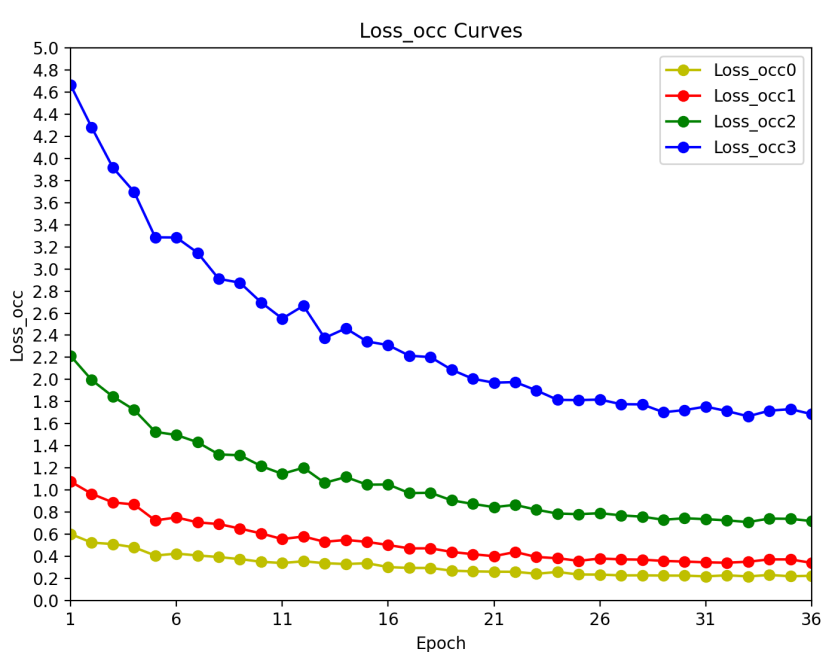}
    \caption{Four layer Loss curve for SurroundOcc training.
    The four curves show a downward trend and converge at last, which indicates that the model has a good effect in the four-layer training.}
    \label{fig:Four Loss}
\end{figure}

\begin{figure}[!ht]  \centering \small
    \includegraphics[width=0.98\linewidth]{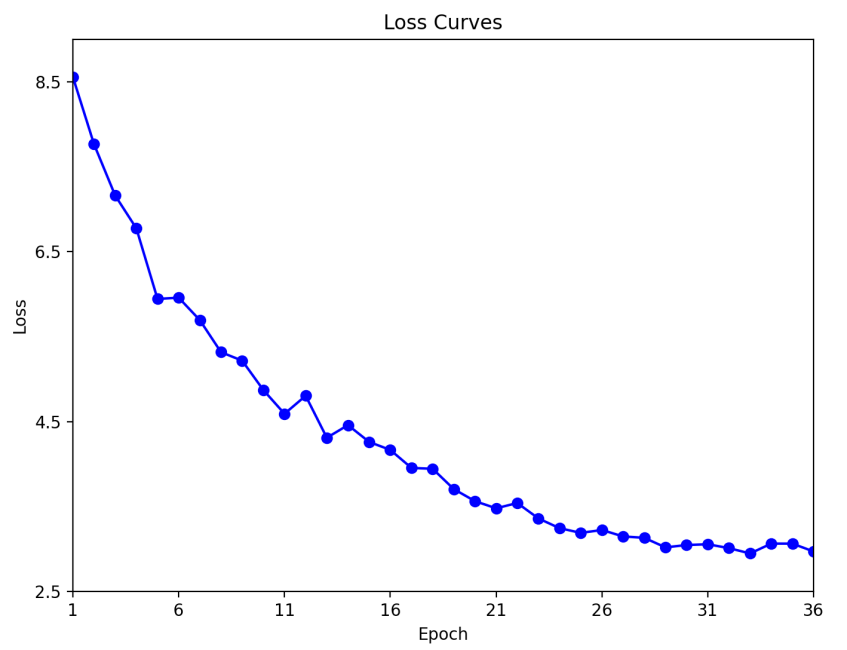}
    \caption{Overall Loss change graph for SurroundOcc training.
    The curve shows a downward trend and finally converges, which indicates that the training of the whole model is normal and effective.}
    \label{fig:whole loss}
\end{figure}

The backbone network of SurroudOcc undergoes supervision at every layer during training, aimed at extracting robust three-dimensional features across both high and low levels for multi-scale occupancy prediction. 
\autoref{fig:Four Loss} illustrates the evolution of loss over the training phases for four distinct layers, while \autoref{fig:whole loss} presents the comprehensive loss trajectory for the entire training process, encapsulating the aggregate performance improvement.

%%%%%%%%%%%%%%%%%%%%%%%%%%%%%%%%%%%%%%%%%%%%%%
\section{Results}
\subsection{Predict underground parking using SurroundOcc pretraining models}

\begin{figure}[!ht]  \centering \small
    \includegraphics[width=0.98\linewidth,trim={80 70 90 50},clip]{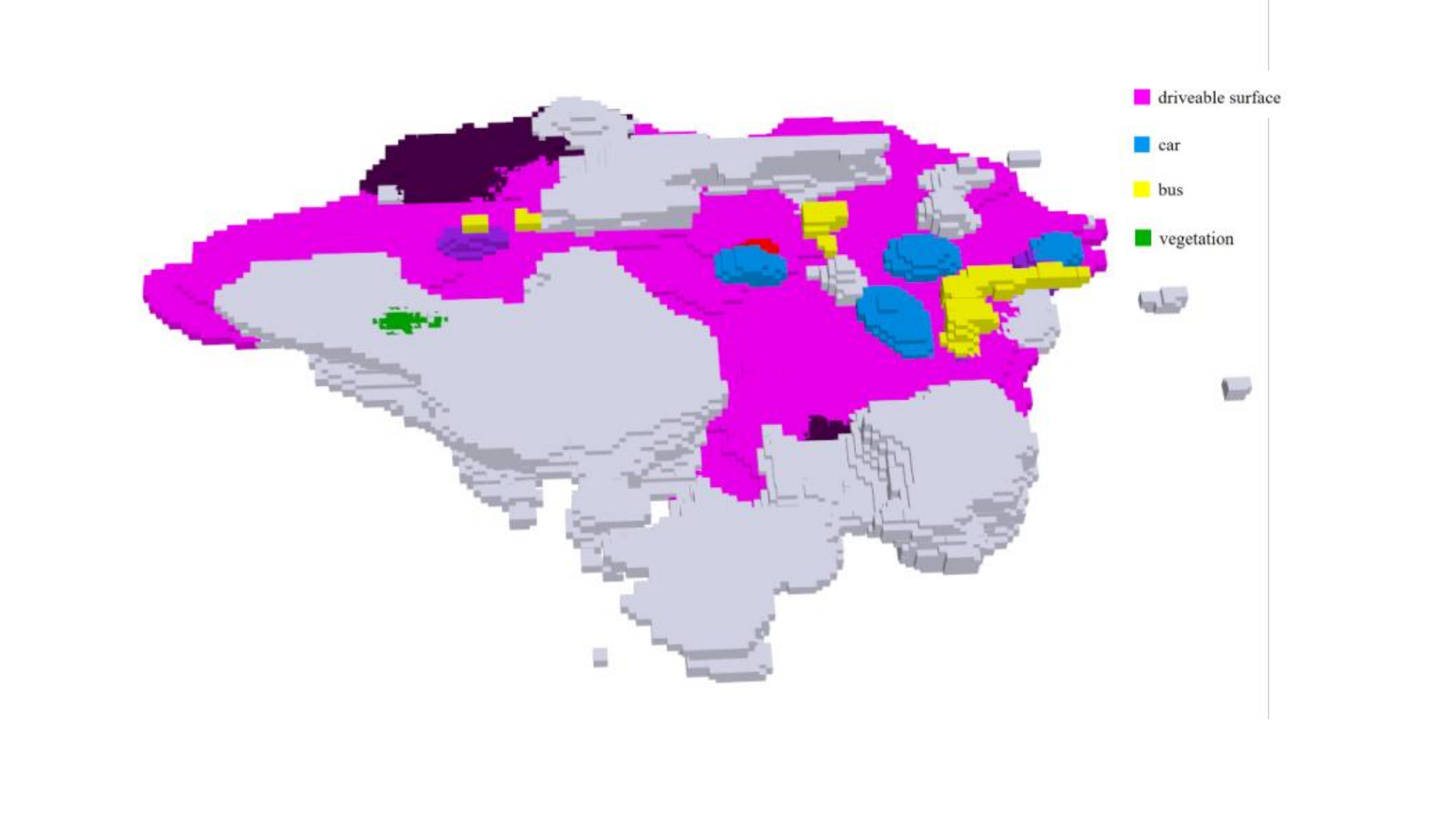}
    \caption{SurroundOcc pretraining model failure prediction for underground parking.
    As a result, there were objects such as buses and plants that would not have appeared in underground parking garages, and the positions and shapes of walls and columns were not accurate.}
    \label{fig:failure prediction}
\end{figure}

Observation of \autoref{fig:failure prediction} reveals inconsistencies between the predicted outcomes and the actual architectural configuration of the underground parking facility. 
Notably, the model fails to discriminate between structural elements such as walls and columns accurately.
Erroneously, vegetation predictions (depicted by the green regions) infiltrate areas designated for walls, which starkly contradicts the typical environment of an underground parking lot.
The pre-training model provided by SurroundOcc is completely ineffective in underground parking.
\subsection{The underground parking lot prediction}
\begin{figure}[!ht]  \centering \small
    \includegraphics[width=0.98\linewidth,trim={20 30 30 30},clip]{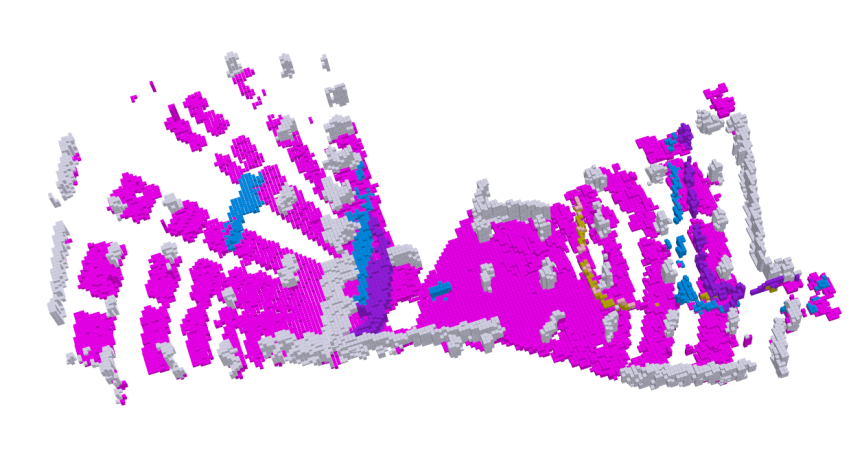}
    \caption{The prediction results of the trained model for the underground parking lot.
    Compared with the corresponding point cloud data map, the positions of the posts, walls, and vehicles are basically correct.}
    \label{fig:prediction result}
\end{figure}

\begin{figure}[!ht]  \centering \small
    \includegraphics[width=0.98\linewidth]{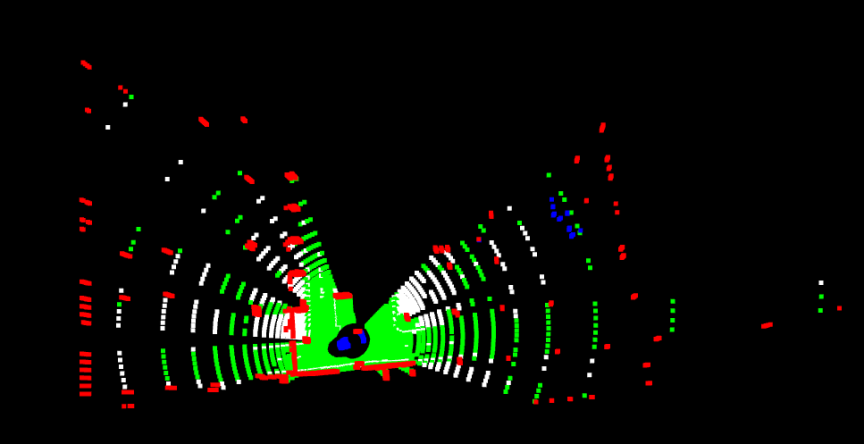}
    \caption{
Visualization point cloud file corresponding to prediction results.}
    \label{fig:predict pcd}
\end{figure}

The model is used to predict the occupancy of the underground parking lot after the training.
The prediction results are shown in \autoref{fig:prediction result}.
The corresponding visual point cloud data is shown in \autoref{fig:predict pcd}.

The model's accuracy was evaluated by calculating the IoU and mIoU values. After calculation, the model's comparison of IoU and mIoU before and after training is shown in \autoref{tab:judge}.
\begin{table}[!ht] \centering \small
  \caption{Comparison of evaluation results before and after training.
  The results show that the accuracy of model prediction is improved.}
  \label{tab:judge}
  \begin{tabular}{c|c|c}   \hline
    state & SC IoU & SSC mIoU  \\    \hline
   before training &  43.8 & 15.5 \\   \hline
    after training & 46.07 & 17.29 \\   \hline
    \end{tabular}
\end{table}
The chart shows that the model's performance has undergone enhancement following training, leading to an improved capacity for occupancy prediction within the network. 
Consequently, the task of perceiving and accurately representing the complex environment of an underground parking lot scenario has been successfully accomplished.
\label{sec:results}

%%%%%%%%%%%%%%%%%%%%%%%%%%%%%%%%%%%%%%%%%%%%%%
\section{Discussion}

\label{sec:discussion}
%%%%%%%%%%%%%%%%%%%%%%%%%%%%%%%%%%%%%%%%%%%%%%

Despite its progress, this study identifies several lingering factors that pose challenges and interfere with the experimental outcomes, with a primary focus on generating true value representations. 
Addressing these issues remains a crucial avenue for future research.
Here are the problems:
\begin{enumerate} [leftmargin=*]
    \item In the generated dense occupancy ground truth, two pillars in an underground parking lot may inadvertently be rendered as a wall.
    It is postulated that this anomaly arises from coordinate transformation errors during the concatenation of radar point clouds. 
    These inaccuracies lead to subsequent pillar point clouds being appended in front of the initial frame's pillar point cloud, visually consolidating them into what appears as a continuous wall surface.
    \item 
    In the synthesized ground truth, instances of semantic interpenetration occur among different objects, such as vehicle voxels infiltrating into the ground. This phenomenon is potentially attributed to the nearest neighbor algorithm, which, when assigning semantics to densely packed voxels, inadvertently allocates the semantics of an object to neighboring voxels that do not genuinely belong to it, thereby causing the object's representation to exceed its actual voxel boundaries.
    \item The confusion of lane markings with wall semantics could stem from discrepancies in the mapping process between the semantic outputs of CARLA's LiDAR scans and the predefined semantics in nuScenes. 
\end{enumerate}

%%%%%%%%%%%%%%%%%%%%%%%%%%%%%%%%%%%%%%%%%%%%%%
\section{Conclusions}
\label{sec:conclusion}
%%%%%%%%%%%%%%%%%%%%%%%%%%%%%%%%%%%%%%%%%%%%%%
\subsection{Work Summary}
This study is dedicated to applying occupancy grid techniques within the intricate environment of underground parking lots for scene prediction, thereby enhancing the environmental perception capabilities of autonomous driving systems and improving driving safety.
The specific research contents and contributions can be summarized into four main aspects:
\begin{enumerate} [leftmargin=*]
    \item By integrating the CARLA Advanced Driving Simulation platform, a comprehensive dataset of underground parking is systematically collected in this study, which covers multi-dimensional information, such as camera images, radar scanning data, fine object labeling, and precise position and attitude information of vehicles and various sensors. 
    It is normalized according to the industry standard nuscenes data format, which provides high-quality basic materials for subsequent deep-learning models.

    \item Based on precise truth generation strategies from the SurroundOcc research team, this study not only replicated this process in depth but optimized the coordinate transformation of the CARLA simulation environment to produce a detailed set of scene occupancy truth values.
    \item Further model optimization \cite{lan2022time,lan2021learning,lan2021learning2} and training was carried out in the SurroundOcc framework.
    This process involves fine-tuning the pre-trained model on a solid basis,
      aiming to make it more suitable for the unique spatial structure and dynamic environmental characteristics of underground parking lots, so as to improve the model's ability to understand and predict complex parking scenarios.
    \item Finally, using a trained and tuned occupancy network model,
    this study realizes occupancy prediction for underground parking lot scenarios. This result verifies the validity of the proposed method.
\end{enumerate}

\subsection{Future Work}
There are still many shortcomings in this study, including the following:
\begin{enumerate}[leftmargin=*]
    \item Only one SurroundOcc model was used to predict underground parking scenes.
    \item The method of generating the true value of occupancy is not completely accurate enough to obtain a completely accurate true value of occupancy.
    \item No real underground parking lot data was used for verification.
\end{enumerate}

In view of the shortcomings in this study, we will continue to complete in the future work:
\begin{enumerate}[leftmargin=*]
    \item Survey more methods of Occupancy network and compare other methods to forecast underground parking lots.
    \item Make improvements on the original method of generating true values to ensure the accuracy of generated true values.
    \item Use real-world parking lot data set to train and verify the model.
    \item Use the predicted results to plan the automatic driving of the vehicle and verify the practical value of the results.
\end{enumerate}

\bibliographystyle{IEEEtran}
\bibliography{bibliography}

\end{document}